\documentclass[lettersize,journal]{IEEEtran}
\usepackage{amsmath,amsfonts}
\usepackage{algorithmic}
\usepackage{algorithm}
\usepackage{array}
\usepackage[caption=false,font=normalsize,labelfont=sf,textfont=sf]{subfig}
\usepackage{textcomp}
\usepackage{stfloats}
\usepackage{url}
\usepackage{verbatim}
\usepackage{graphicx}
\usepackage{cite}
\usepackage{CJKutf8}         
\usepackage[hidelinks]{hyperref}
\usepackage{arydshln}
\usepackage{multirow}
\usepackage{booktabs}
\usepackage{tcolorbox}
\usepackage{xcolor}         
\usepackage{tabularx}
\usepackage{array}
\definecolor{LimeGreen}{rgb}{0.2, 0.8, 0.2}
\definecolor{Red}{rgb}{1, 0, 0}
\newcolumntype{C}[1]{>{\centering\arraybackslash}p{#1}}

\begin{document}

\title{Aligning LLM Uncertainty with Human Disagreement in Subjectivity Analysis}

\author{Junyu~Lu,
        Deyi~Ji,
        Xuanyi~Liu,
        Lanyun~Zhu,
        Bo~Xu,
        Liang~Yang,  
        Xian-Sheng~Hua,
        Hongfei~Lin%
\thanks{Junyu Lu, Bo Xu, Liang Yang, and Hongfei Lin are with the School of Computer Science and Technology, Dalian University of Technology, China. Deyi Ji is with Tencent. Xuanyi Liu is with Peking University. Lanyun Zhu and Xian-Sheng Hua are with Tongji University.}
}

\markboth{Journal of \LaTeX\ Class Files,~Vol.~14, No.~8, August~2021}%
{Shell \MakeLowercase{\textit{et al.}}: A Sample Article Using IEEEtran.cls for IEEE Journals}


\maketitle

\begin{abstract}
Large language models for subjectivity analysis are typically trained with aggregated labels, which compress variations in human judgment into a single supervision signal.
This paradigm overlooks the intrinsic uncertainty of low-agreement samples and often induces overconfident predictions, undermining reliability and generalization in complex subjective settings.
In this work, we advocate uncertainty-aware subjectivity analysis, where models are expected to make predictions while expressing uncertainty that reflects human disagreement.
To operationalize this perspective, we propose a two-phase \textbf{D}isagreement \textbf{P}erception and \textbf{U}ncertainty \textbf{A}lignment \textbf{(DPUA)} framework.
Specifically, DPUA jointly models label prediction, rationale generation, and uncertainty expression under an uncertainty-aware setting.
In the disagreement perception phase, adaptive decoupled learning enhances the model's sensitivity to disagreement signals while preserving task performance.
In the uncertainty alignment phase, GRPO-based reward optimization further improves uncertainty-aware reasoning and aligns the model's confidence expression with the human disagreement distribution.
Experiments on three subjectivity analysis tasks show that DPUA preserves task performance while better aligning model uncertainty with human disagreement, mitigating overconfidence on boundary samples, and improving out-of-distribution generalization.

\textcolor{red}{\textit{Disclaimer}:\textit{ The paper contains content that may be profane, vulgar, or offensive.}}
\end{abstract}


\section{Introduction}

In recent years, large language models (LLMs) have achieved remarkable progress in natural language processing, driven by their extensive background knowledge and strong generative capabilities \cite{DBLP:conf/nips/KojimaGRMI22, DBLP:journals/corr/abs-2303-08774, DBLP:journals/jmlr/ChowdheryNDBMRBCSGSSTMRBTSPRDHPBAI23}.
The dominant training paradigm typically relies on standard labels as supervision signals \cite{DBLP:journals/corr/abs-2502-04194}.
This paradigm has been highly effective for objective tasks such as mathematical reasoning \cite{DBLP:conf/iclr/Zhou0NLWWHWH25} and factual question answering \cite{DBLP:conf/iclr/HuCLGWYG24}, where correctness criteria are relatively clear and stable.

However, this paradigm becomes less reliable for subjectivity analysis tasks, such as sentiment classification, sarcasm detection, and offensiveness detection.
Unlike objective tasks, subjectivity analysis often involves fuzzy semantic boundaries and diverse human interpretations \cite{DBLP:conf/www/AroyoDTRR19, DBLP:journals/jair/UmaFHPPP21}.
As a result, different humans may reasonably assign different labels to the same input.
Such disagreement is not merely annotation noise; rather, it often reflects intrinsic sample uncertainty and variation in human cognition \cite{DBLP:conf/aiia/Basile20, basile-etal-2021-need}.
When these differences are compressed into a single aggregated label, conventional supervision overlooks the uncertainty of low-agreement samples, which may induce overly rigid decision boundaries and overconfident predictions \cite{uma-etal-2021-semeval, DBLP:conf/semeval/LeonardelliAABF23}.
This undermines model reliability and generalization in complex subjective scenarios \cite{DBLP:journals/tacl/DavaniDP22, DBLP:conf/acl/LuMWXL00L25}.

\begin{figure}[t]
    \centering
    \includegraphics[width=1.0\linewidth]{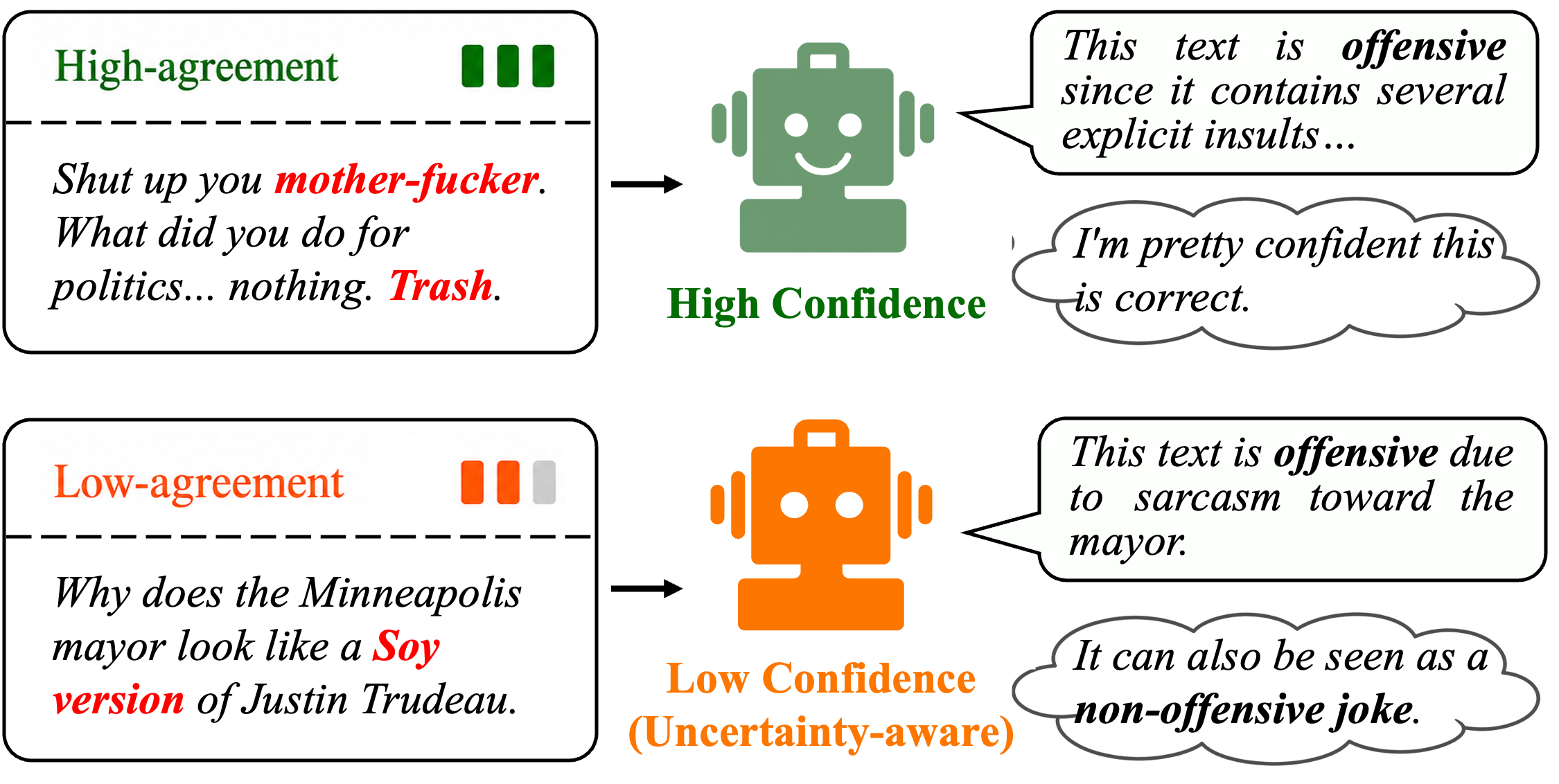}
    \caption{Illustration of uncertainty-aware subjectivity analysis in offensive language detection. High-agreement samples support confident predictions, whereas low-agreement samples require more cautious uncertainty expression due to multiple plausible interpretations.}   
\label{fig:intro}
\end{figure}

To address this limitation, we advocate uncertainty-aware subjectivity analysis.
Instead of treating the aggregated label as the only learning target, this setting requires models to make task decisions while expressing uncertainty that reflects human disagreement.
In other words, an ideal model should not only answer \textit{``what the label is''}, but also indicate \textit{``how uncertain the case is''}.
Such uncertainty awareness should be grounded in task evidence and cues that may lead to human disagreement.
As illustrated in Figure~\ref{fig:intro}, in offensive language detection, a model should be confident on high-agreement samples with explicit profanity or insults, while being more cautious on low-agreement samples that allow multiple plausible interpretations.
This ability to align confidence with human disagreement is important for mitigating overconfidence on boundary samples and supporting more reliable decisions in subjective analysis tasks.

To this end, we propose a two-phase \textbf{D}isagreement \textbf{P}erception and \textbf{U}ncertainty \textbf{A}lignment framework (\textbf{DPUA}) for subjectivity analysis.
DPUA jointly generates a task label, a rationale, and a confidence score, where the rationale captures task evidence and disagreement-related cues, and the confidence score reflects the model's uncertainty.
The first phase performs disagreement perception through adaptive decoupled learning, which adjusts supervision according to human agreement and encourages the model to learn stable decision patterns from high-agreement samples while attending to disagreement signals in low-agreement cases.
The second phase performs uncertainty alignment with Group Relative Policy Optimization (GRPO)~\cite{DBLP:journals/corr/abs-2402-03300}, further improving uncertainty-aware reasoning and aligning confidence expression with the human disagreement distribution.

We conduct a systematic evaluation of DPUA on three subjectivity analysis tasks.
Experimental results show that DPUA preserves core task performance while better aligning model uncertainty with human disagreement, effectively mitigating overconfidence on boundary samples.
Furthermore, under out-of-distribution settings, DPUA shows stronger generalization across different subjectivity analysis tasks.
The contributions of this paper are summarized as follows:

\begin{itemize}
\item We advocate uncertainty-aware subjectivity analysis, where models are expected to predict task labels, identify disagreement-related cues, and express calibrated uncertainty beyond fitting a single aggregated label.
\item We propose DPUA, a two-phase Disagreement Perception and Uncertainty Alignment framework that enables LLMs to identify disagreement signals and express uncertainty in accordance with human disagreement.
\item Experiments on three subjectivity analysis tasks demonstrate that DPUA preserves task performance while improving uncertainty alignment, mitigating overconfidence on boundary samples, and enhancing out-of-distribution generalization.
\end{itemize}

\section{Related Work}

\subsection{Subjectivity Analysis.}
Subjectivity analysis tasks pose a particular challenge for LLMs \cite{DBLP:conf/acl/LuMWXL00L25}, as they require models to interpret texts whose meanings are often context-dependent and open to multiple plausible readings. 
Such tasks involve inherently complex judgment processes and substantial human disagreement, particularly on boundary samples where semantic cues are ambiguous or socially sensitive \cite{uma-etal-2021-semeval, basile-etal-2021-need}.
Early studies typically collapsed such disagreement into a single supervision signal through majority voting \cite{DBLP:conf/aaai/MathewSYBG021,DBLP:conf/icwsm/DavidsonWMW17}. 
Recently, several studies have argued that this binary classification formulation relies on an idealized assumption \cite{basile-etal-2021-need, DBLP:conf/aiia/Basile20,plank-2022-problem}. 
In response, recent studies have begun to explore how human disagreement information can be incorporated into model training \cite{DBLP:conf/naacl/FornaciariUPPHP21, DBLP:conf/acl/BaumlerSD23, weerasooriya-etal-2023-disagreement, DBLP:conf/emnlp/ParappanH25}.
However, most existing studies still focus primarily on improving predictive performance, while research on enabling LLMs to explicitly represent and align with human disagreement remains limited.

\subsection{LLM Uncertainty Calibration.}
Model uncertainty in LLMs has received increasing attention, as it is closely related to the reliability and trustworthiness of model outputs \cite{DBLP:journals/corr/abs-2410-15326, DBLP:journals/csur/ShorinwaMLRM26}. 
Existing studies have explored various ways to estimate or express uncertainty \cite{DBLP:journals/ijmms/XuSL25, DBLP:journals/tmlr/LinHE22, zhang2026tokurtokenleveluncertaintyestimation, DBLP:conf/acl/FadeevaRSPLMTKP24}, aiming to help models better reflect the reliability of their outputs through confidence estimates.
These methods have shown promising results in tasks where correctness can be more clearly defined \cite{DBLP:conf/iclr/0002WSLCNCZ23, DBLP:conf/acl/Chen024, DBLP:conf/aaai/0001SM0ETASC25}, effectively improving model reliability.
Most existing work, however, is developed for objective settings and evaluates uncertainty mainly with respect to prediction correctness \cite{DBLP:journals/corr/abs-2410-15326, DBLP:journals/csur/ShorinwaMLRM26}. 
This formulation is less suitable for subjectivity analysis tasks, where uncertainty often stems from uncertainty in the sample and variation in human cognition rather than model error alone \cite{DBLP:conf/eacl/SandriLTJ23, DBLP:conf/emnlp/DengZ0W0M23}. 
As a result, methods designed for objective tasks cannot be directly transferred to this setting \cite{weerasooriya-etal-2023-disagreement, DBLP:conf/acl/LuMWXL00L25}. 
How to model uncertainty in a way that better reflects human disagreement remains underexplored.

\section{Methodology}

\begin{figure*}[t]
    \centering
    \includegraphics[width=1.0\linewidth]{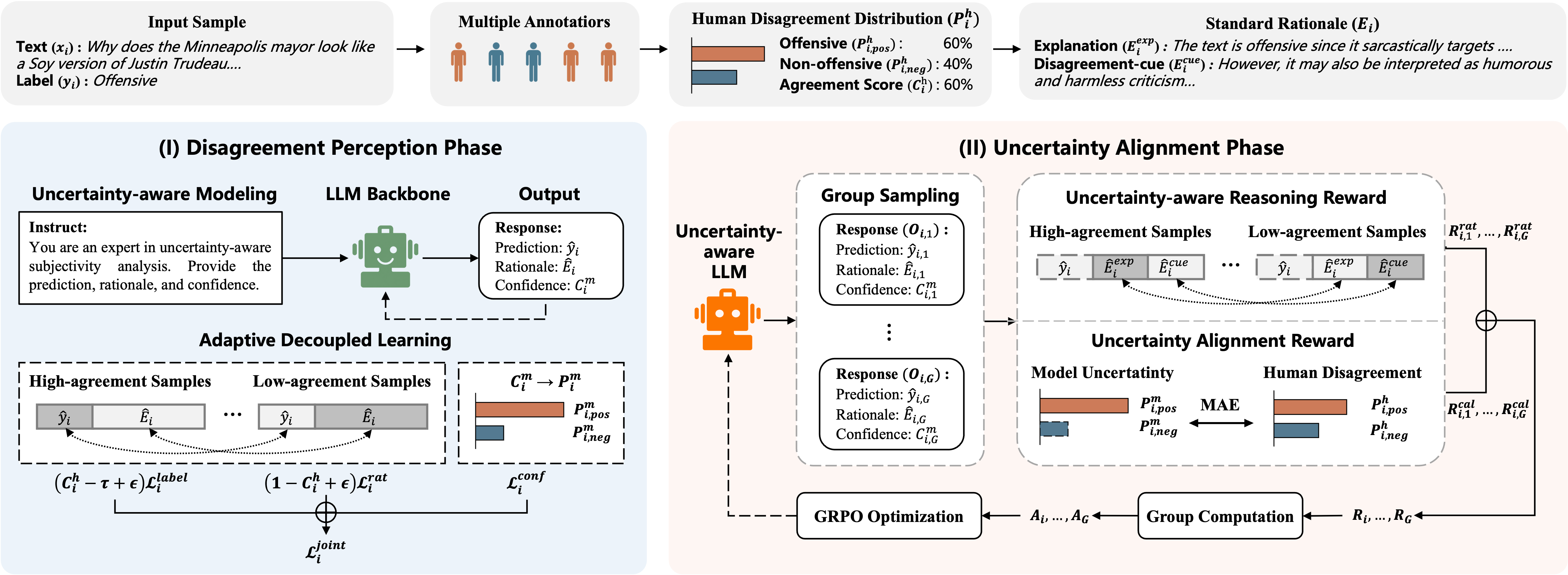}
    \caption{Illustration of Disagreement Perception and Uncertainty Alignment (DPUA) framework. The model jointly performs label prediction, rationale generation, and confidence estimation. Disagreement perception enhances sensitivity to disagreement signals, while uncertainty alignment calibrates confidence to reflect human disagreement.} 
    \vspace{-0.05in}
\label{fig:framework}
\end{figure*}

\subsection{Overview}

To operationalize uncertainty-aware subjectivity analysis, we propose a two-phase \textbf{D}isagreement \textbf{P}erception and \textbf{U}ncertainty \textbf{A}lignment framework (\textbf{DPUA}).
DPUA enables LLMs to make task predictions while expressing uncertainty that reflects human disagreement.
Specifically, we introduce an uncertainty-aware modeling setting in which the model jointly performs label prediction, rationale generation, and uncertainty expression.
DPUA consists of two phases: \textit{disagreement perception} and \textit{uncertainty alignment}.
In the disagreement perception phase, adaptive decoupled learning enhances the model's sensitivity to disagreement signals while preserving task performance.
In the uncertainty alignment phase, GRPO-based reward optimization further improves uncertainty-aware reasoning and aligns the model's confidence expression with the human disagreement distribution.
Figure~\ref{fig:framework} illustrates the overall framework of DPUA.
In the following sections, we first present the task formulation and uncertainty-aware modeling setting, and then describe the two phases in detail.

\subsection{Task Formulation}

Due to the limited availability of subjectivity analysis datasets with non-aggregated annotations, we focus on three binary subjectivity analysis tasks: sarcasm detection, sentiment analysis, and offensiveness detection.
Given an input sample $x_i$, the standard objective is to predict its majority hard label $y_i \in \{\mathrm{pos}, \mathrm{neg}\}$, where $\mathrm{pos}$ and $\mathrm{neg}$ denote the task-specific positive and negative classes, respectively.
In this setting, the model produces a label prediction $\hat{y}_i$, and the learning objective is commonly formulated as maximizing the likelihood of the target label:
\begin{equation}
\mathcal{L}_i^{\mathrm{label}}
=
-\log p(y_i \mid x_i).
\end{equation}
This conventional formulation treats the majority label as the only supervision signal and implicitly assumes that each sample has a single reliable ground truth.
However, in subjectivity analysis tasks, samples often involve semantic ambiguity and diverse human interpretations, making it insufficient to rely solely on the aggregated hard label.

\subsection{Uncertainty-aware Modeling}

To model the intrinsic uncertainty in subjectivity analysis, we further retain the human disagreement distribution $\mathbf{p}_i^{h}$ for each sample:
\begin{equation}
\mathbf{p}_i^{h} = [p^{h}_{i,\mathrm{pos}},\, p^{h}_{i,\mathrm{neg}}],
\end{equation}
where $p^{h}_{i,\mathrm{pos}}$ and $p^{h}_{i,\mathrm{neg}}$ denote the proportions of annotators assigning the task-specific positive and negative labels, respectively.
Based on this distribution, we define the human agreement score as
\begin{equation}
C_i^{h} = \max\left(p^{h}_{i,\mathrm{pos}},\, p^{h}_{i,\mathrm{neg}}\right)\in[0.5,1],
\end{equation}
which measures the degree of annotator agreement, with lower values indicating stronger disagreement.

Given an input sample $x_i$, the model is required not only to predict its label, but also to generate a rationale and express its uncertainty through a confidence score.
Since existing datasets do not provide human-written rationales, we use the majority label and human disagreement distribution to guide GPT-4o in distilling reference rationales, which are then manually verified for reliability and denoted as $E_i$.
Each rationale consists of two components: label justification and disagreement-cue identification.
Accordingly, we represent the reference rationale and the generated rationale as
\begin{equation}
E_i = \left(E_i^{\mathrm{lab}},\, E_i^{\mathrm{cue}}\right), \qquad
\hat{E}_i = \left(\hat{E}_i^{\mathrm{lab}},\, \hat{E}_i^{\mathrm{cue}}\right),
\end{equation}
respectively.
The model output is then defined as $(\hat{y}_i, \hat{E}_i, C_i^{m})$, where $\hat{y}_i$ is the predicted label, $\hat{E}_i$ is the generated rationale, and $C_i^{m}\in(0.5,1]$ denotes the confidence score assigned to the predicted label.
Values closer to $0.5$ indicate greater uncertainty, while values closer to $1$ indicate higher certainty.

To support the subsequent two-phase optimization, we formulate a unified autoregressive objective over three output components: label prediction, rationale generation, and confidence estimation.
Specifically, for each sample $x_i$, the target output is divided into three disjoint segments corresponding to $\hat{y}_i$, $\hat{E}_i$, and $C_i^{m}$, with token index sets denoted by $\mathcal{I}_i^{\mathrm{label}}$, $\mathcal{I}_i^{\mathrm{rat}}$, and $\mathcal{I}_i^{\mathrm{conf}}$, respectively.
Here, the rationale segment $\mathcal{I}_i^{\mathrm{rat}}$ covers both the label-justification and disagreement-cue components.
The segment-level loss for each part is defined as
\begin{equation}
\mathcal{L}_i^{k}
=
\frac{1}{|\mathcal{I}_i^{k}|}
\sum_{t\in \mathcal{I}_i^{k}}
-\log p(s_{i,t}\mid x_i, s_{i,<t}),
\end{equation}
where $k\in\{\mathrm{label}, \mathrm{rat}, \mathrm{conf}\}$.

Under this formulation, the model is expected to predict the majority label while generating rationales and confidence scores that reflect the intrinsic uncertainty induced by human disagreement.
This formulation provides a basis for the subsequent disagreement perception and uncertainty alignment phases.
The detailed instruction template is shown as follows:

\begin{tcolorbox}[colback=gray!3, colframe=black, arc=2mm, boxrule=0.25mm, boxsep=0mm]
\small
\textbf{\textit{Instruction}}:\\
You are an expert in uncertainty-aware subjectivity analysis, capable of making judgments and expressing uncertainty in a human-aligned manner.\\

\textbf{\textit{Task}}:\\
- Task Definition: \texttt{<task definition>}\\
- Label Space: \texttt{<pos, neg>}\\
- Given the input text,\\
1) Determine whether the input is more appropriately labeled as \texttt{<pos>} or \texttt{<neg>}.\\
2) Provide a rationale based on the contextual cues in the text, and analyze potential uncertainty in terms of semantic ambiguity or possible annotator disagreement, preferably using transitional phrases such as \textit{although} or \textit{however}.\\
3) Provide a confidence score for this judgment based on the clarity and completeness of the evidence. The score must be a floating-point number in the range $(0.5, 1.0]$.\\

\textbf{\textit{Output Format}}:\\
- Prediction: \texttt{<prediction label>}\\
- Rationale: \texttt{<rationale for your prediction>}\\
- Confidence: \texttt{<a float in (0.5, 1.0]>}\\

\textbf{\textit{Input}}:\\
\texttt{<input text>}
\end{tcolorbox}

\subsection{Disagreement Perception Phase}

To enhance the model's ability to identify disagreement signals, we introduce the disagreement perception phase as the first phase of DPUA, implemented through supervised fine-tuning (SFT).
This phase adopts an adaptive decoupled learning strategy that selectively reweights label and rationale learning according to the human agreement score.
In this way, the model learns reliable task decisions from high-agreement samples while becoming more sensitive to ambiguity and disagreement cues in low-agreement samples.

\textbf{Adaptive Decoupled Learning.}
Built on the standard segment-level objective in Eq.~(5), adaptive decoupled learning adjusts the relative emphasis on label prediction, rationale generation, and confidence learning according to the human agreement score $C_i^h$.
High-agreement samples provide more reliable supervision for decision learning, while low-agreement samples are assigned larger weights in rationale learning to help the model capture disagreement-related cues.
Confidence learning is optimized across all samples to encourage consistent uncertainty expression.

Formally, the objective of the disagreement perception phase is defined as
\begin{equation}
\small
\mathcal{L}_{\mathrm{joint}}
=
\sum_{i=1}^{N}
\left[
\big(C_i^h-\tau+\epsilon\big)\mathcal{L}_i^{\mathrm{label}}
+
\alpha\big(1-C_i^h+\epsilon\big)\mathcal{L}_i^{\mathrm{rat}}
+
\mathcal{L}_i^{\mathrm{conf}}
\right],
\end{equation}
where $N$ denotes the number of samples in the current mini-batch, $\tau=0.5$ is the minimum possible agreement score in binary classification, $\alpha$ controls the overall strength of rationale learning to balance its influence against label prediction and confidence learning, and $\epsilon$ is a smoothing term that preserves a minimum level of label and rationale supervision.

In this objective, the agreement-based weighting term strengthens label supervision for high-agreement samples, thereby preserving task performance under uncertainty-aware learning.
The rationale weighting term increases the contribution of low-agreement samples, encouraging the model to learn disagreement-related evidence.
The confidence term is not explicitly modulated by $C_i^h$, since uncertainty expression is treated as a global behavioral property learned across both high- and low-agreement samples.

Through this strategy, the model focuses more on stable decision patterns in high-agreement cases while allocating more learning capacity to disagreement signals in low-agreement cases, establishing a suitable foundation for the subsequent uncertainty alignment phase.

\subsection{Uncertainty Alignment Phase}

While the disagreement perception phase helps the model recognize disagreement signals through supervised learning, token-level imitation alone does not explicitly reward outputs that reason about uncertainty or align confidence scores with human disagreement.
To this end, we introduce an uncertainty alignment phase based on GRPO \cite{DBLP:journals/corr/abs-2402-03300}, which supports reasoning-oriented optimization through relative comparisons among multiple sampled outputs.
Building on the disagreement signals learned in the previous phase, this phase further refines rationale generation and aligns the model's uncertainty expression with the human disagreement distribution.

\textbf{Uncertainty-aware Reasoning Reward.}
We define an uncertainty-aware reasoning reward by evaluating the two components of the generated rationale: label justification and disagreement-cue identification.
Specifically, we employ a judge model based on GPT-4o-mini to compare the generated rationale components $\hat{E}_i^{\mathrm{lab}}$ and $\hat{E}_i^{\mathrm{cue}}$ with their reference counterparts $E_i^{\mathrm{lab}}$ and $E_i^{\mathrm{cue}}$.
For each component, the judge assigns a score on a 3-point Likert scale, where 1, 2, and 3 denote weak, moderate, and strong alignment, respectively.
These Likert scores are linearly normalized to $[0,1]$ before reward computation, and the normalized scores are denoted as $s_i^{\mathrm{lab}}$ and $s_i^{\mathrm{cue}}$.

We further use the human agreement score $C_i^h$ to dynamically balance these two aspects of rationale evaluation.
For high-agreement samples, the reward places greater emphasis on label justification; for low-agreement samples, it assigns more weight to disagreement-related cues.
Accordingly, we define the uncertainty-aware reasoning reward as
\begin{equation}
R_i^{\mathrm{rat}}
=
\frac{
(C_i^h-\tau+\epsilon) \, s_i^{\mathrm{lab}}
+
(1-C_i^h+\epsilon)\, s_i^{\mathrm{cue}}
}{
1-\tau+2\epsilon
} \in[0, 1].
\end{equation}
This design encourages the model to adapt its rationale generation to the agreement structure of each sample.
For high-agreement cases, the reward favors accurate label justification, whereas for low-agreement cases, it places greater emphasis on identifying disagreement-related cues.

\textbf{Uncertainty Calibration Reward.}
Beyond rationale quality, we further optimize whether the model expresses uncertainty in a way that is calibrated to human disagreement.
For each input sample $x_i$, the model predicts a label $\hat{y}_i$ and assigns a confidence score $C_i^m$ to this prediction.
To compare this model-side confidence with human disagreement, we first convert the prediction-confidence pair into a class-wise uncertainty distribution in the binary setting:
\begin{equation}
\mathbf{p}^{m}_{i} =
\begin{cases}
[C_i^m,\, 1 - C_i^m], & \text{if } \hat{y}_i = \mathrm{pos},\\
[1 - C_i^m,\, C_i^m], & \text{if } \hat{y}_i = \mathrm{neg}.
\end{cases}
\end{equation}
Here, $\mathbf{p}^{m}_{i}$ represents the model-implied distribution over the positive and negative classes.

On the human side, we directly use the empirical human disagreement distribution $\mathbf{p}^{h}_{i}$ obtained from multiple annotations.
We then measure the calibration discrepancy between $\mathbf{p}^{m}_{i}$ and $\mathbf{p}^{h}_{i}$ using Mean Absolute Error (MAE):
\begin{equation}
\mathrm{MAE}_i
=
\frac{1}{2}
\sum_{c\in\{\mathrm{pos},\mathrm{neg}\}}
\left| p_{i,c}^{m} - p_{i,c}^{h} \right|.
\end{equation}
A smaller MAE indicates that the model's uncertainty expression is more consistent with the human disagreement distribution.
Therefore, we define the uncertainty calibration reward as
\begin{equation}
R_i^{\mathrm{cal}} = 1-\mathrm{MAE}_i \in[0, 1].
\end{equation}

This reward encourages the model to calibrate its confidence according to the degree of human disagreement.
For high-agreement samples, it rewards confident predictions that match the dominant human judgment; for low-agreement samples, it discourages overconfident outputs and promotes uncertainty expression closer to human annotation variation.

\textbf{GRPO Optimization.}
For each input sample $x_i$, we generate $G$ sampled outputs, where $g\in\{1,\dots,G\}$ indexes an output in the group.
Following GRPO, these outputs are compared through relative rewards, enabling the model to optimize reasoning quality and uncertainty expression via group-wise preference learning.
We combine the two reward components to obtain the overall reward for each sampled output:
\begin{equation}
R_{i,g}
=
R_{i,g}^{\mathrm{rat}} + R_{i,g}^{\mathrm{cal}}.
\end{equation}

Given the reward scores of the $G$ sampled outputs for input $x_i$, we compute the group-relative advantage by normalizing rewards within the group:
\begin{equation}
A_{i,g}
=
\frac{R_{i,g} - \mathrm{mean}(\{R_{i,1},\dots,R_{i,G}\})}
{\mathrm{std}(\{R_{i,1},\dots,R_{i,G}\}) + \epsilon_{\mathrm{adv}}}.
\end{equation}
Here, $A_{i,g}$ measures the relative quality of the $g$-th sampled output within the group, and $\epsilon_{\mathrm{adv}}$ is a small constant for numerical stability.

Based on these advantages, we further refine the model learned in the disagreement perception phase, denoted as $\pi_\theta$, using GRPO:
\begin{equation}
\mathcal{L}_{\mathrm{align}}
=
-
\frac{1}{N}
\sum_{i=1}^{N}
\frac{1}{G}
\sum_{g=1}^{G}
A_{i,g}
\sum_{t=1}^{|o_{i,g}|}
\log \pi_{\theta}(o_{i,g,t}\mid x_i, o_{i,g,<t}).
\end{equation}

Through GRPO-based optimization, the uncertainty alignment phase further refines the model beyond supervised learning by enhancing uncertainty-aware reasoning and aligning its uncertainty expression with the underlying human disagreement distribution.
As a result, the model becomes better able to express uncertainty in a manner consistent with the degree of human disagreement in subjectivity analysis tasks.

\subsection{Implementation Details}
To ensure the quality of training signals, we manually verify distilled rationales and LLM-as-Judge evaluation. 
For GPT-4o-generated rationales, we examine 500 samples and find that most rationales correctly support the majority label and capture plausible disagreement cues. 
For LLM-as-Judge evaluation in the uncertainty-aware reasoning reward, we manually annotate another 500 samples using the same 3-point Likert scale. 
The exact-match agreement between human and judge scores reaches 94.5\%, suggesting that the judge model provides consistent reward signals.

\section{Experiment}

\begin{table*}[t]
  \centering
  \setlength{\tabcolsep}{5pt}
  \caption{Statistics of the datasets used. Here, \#Anno. refers to the number of annotators, $C^h$ denotes the human consensus score of samples, and Avg. L denotes the average text length in words.}
  \begin{tabular}{C{4cm}C{1.2cm}C{1.2cm}C{1.2cm}C{1.2cm}C{1.2cm}C{1.2cm}C{1.2cm}C{1.2cm}}
    \toprule
    Dataset (\textit{Pos. / Neg.}) & \#Pos. & \#Neg. & \#Train & \#Test & \#Total & \#Anno. & Avg. $C^h$ & Avg. L \\
    \midrule
    CSC (\textit{Sarcastic / Normal}) & 1,490 & 4,422 & 5,240 & 672  & 5,912 & 4-6 & 0.76 & 53.02 \\
    GOE (\textit{Positive / Negative}) & 1,035 & 1,040 & 2,075 & 519  & 2,594 & 3-5 & 0.75 & 14.60 \\
    MD  (\textit{Offensive / Non-offensive}) & 2,980 & 6,669 & 6,592 & 3,057 & 9,649 & 5 & 0.83 & 23.06 \\
    \bottomrule
  \end{tabular}
  \label{statistics}
\end{table*}

\subsection{Experiment Setup}

\textbf{Datasets.} 
We evaluate DPUA on three subjectivity analysis tasks: sarcasm detection, offensiveness detection, and sentiment analysis. 
\begin{itemize}

\item For sarcasm detection, we use Conversational Sarcasm Corpus (CSC) \cite{DBLP:conf/naacl/JangF24}, a context-aware dataset that requires the model to determine whether a response is sarcastic given the dialogue context and the reply content. 
\item For sentiment analysis, we utilize GoEmotions (GOE) \cite{DBLP:conf/acl/DemszkyMKCNR20} to identify the emotional polarity expressed in text by distinguishing between positive and negative emotion categories.
\item For offensiveness detection, we employ MD-Agreement (MD) \cite{DBLP:conf/emnlp/LeonardelliMAGT21}, which aims to identify whether an online post is offensive, with particular emphasis on more implicit forms of aggression. 
\end{itemize}
All datasets provide non-aggregated annotations and have been widely used in prior research on human disagreement \cite{uma-etal-2021-semeval, DBLP:conf/semeval/LeonardelliAABF23, DBLP:journals/corr/abs-2510-08460}. 
After quality control, the remaining disagreement is generally treated in prior work as a useful proxy for intrinsic ambiguity in human interpretation, rather than being simply discarded as annotation error or random noise.
We directly adopt the training and test split provided in these datasets.
The statistics of datasets are reported in Table~\ref{statistics}.

\textbf{Metrics.}
We evaluate model performance from two complementary aspects: label prediction and uncertainty alignment.
For label prediction, we use Accuracy (Acc.) and macro-F1 score (F1) to assess whether the model correctly predicts task labels.
For uncertainty alignment, we use Mean Absolute Error (MAE) and Pearson correlation coefficient (Coef.) to evaluate whether the model uncertainty distribution (Eq. (8)) is aligned with the human disagreement distribution (Eq. (2)). 
Both distributions are defined over the task label space and thus preserve task-label polarity.
MAE and Coef. are then computed between these two distributions, where MAE measures their absolute discrepancy and Coef. measures the consistency of their variation trends.


\textbf{Experimental Settings.} 
In the disagreement perception phase, we perform SFT for 3 epochs with a learning rate of $1\times10^{-4}$. 
In the uncertainty alignment phase, we employ GRPO for 1 epoch with a learning rate of $1\times10^{-6}$. 
To improve training efficiency, we adopt LoRA (Low-Rank Adaptation) with a rank of 16 and a dropout rate of 0.05. 
The rationale learning weight $\alpha$, $\epsilon$, and $\epsilon_{\mathrm{adv}}$ are all set to 0.1.
The maximum rationale length is set to 100 tokens. 
Models are trained on the training set and evaluated on the test set. 
All results are averaged over three runs. 
Experiments are conducted on four NVIDIA H20 GPUs. 

\textbf{Baselines.}
For a comprehensive evaluation, we use Qwen-3-Instruct (8B) and LLaMa-3.1-Instruct (8B) as the backbone models.
We compare DPUA with four representative paradigms: zero-shot inference, few-shot prompting, supervised fine-tuning (SFT), and GRPO-based optimization.
For few-shot prompting, we randomly select one positive and one negative example from each task as demonstrations.

For both SFT and GRPO, we consider two training settings.
The standard setting uses only the majority hard label as supervision: SFT trains the model for label prediction, and GRPO further optimizes the standard SFT checkpoint with a vanilla accuracy-based reward, where correct predictions receive +1 and incorrect predictions receive -1.
The uncertainty-aware setting, denoted with $^*$, augments the training target with rationale generation and confidence estimation following Eq.~(5).
Accordingly, SFT$^*$ is trained with the unified autoregressive objective, and GRPO$^*$ further optimizes the SFT$^*$ checkpoint with the same accuracy-based reward.
At inference time, all methods are required to output both the predicted label and the confidence score.
For fairness, the hyperparameter settings of SFT and GRPO are kept identical to those used in DPUA.

\begin{table*}[t]
  \centering
    \caption{Overall performance comparison on three subjectivity analysis datasets. Results are reported in terms of Acc., F1, MAE, and Coef.; $^*$ denotes baselines augmented with uncertainty-aware modeling. \textbf{Bold} and \underline{underlined} scores indicate the best and second-best results, respectively.}
  \setlength{\tabcolsep}{7pt}
  \renewcommand{\arraystretch}{1.08}
  \begin{tabular}{
    >{\raggedright\arraybackslash}m{1.5cm}
    >{\centering\arraybackslash}m{0.85cm}
    >{\centering\arraybackslash}m{0.85cm}
    >{\centering\arraybackslash}m{0.85cm}
    >{\centering\arraybackslash}m{0.85cm}
    >{\centering\arraybackslash}m{0.85cm}
    >{\centering\arraybackslash}m{0.85cm}
    >{\centering\arraybackslash}m{0.85cm}
    >{\centering\arraybackslash}m{0.85cm}
    >{\centering\arraybackslash}m{0.85cm}
    >{\centering\arraybackslash}m{0.85cm}
    >{\centering\arraybackslash}m{0.85cm}
    >{\centering\arraybackslash}m{0.85cm}
  }
    \toprule
    \multirow{2}{*}{Methods}
          & \multicolumn{4}{c}{CSC}
          & \multicolumn{4}{c}{GOE}
          & \multicolumn{4}{c}{MD} \\
    \cmidrule(lr){2-5}
    \cmidrule(lr){6-9}
    \cmidrule(lr){10-13}
          & Acc.  & F1    & MAE   & Coef.
          & Acc.  & F1    & MAE   & Coef.
          & Acc.  & F1    & MAE   & Coef. \\
    \midrule
    \multicolumn{13}{c}{Qwen-3-Instruct (8B)} \\
    \midrule
    zero-shot      & 0.4851 & 0.4850 & 0.4231 & 0.3439 & 0.7694 & 0.7609 & 0.2350 & 0.6571 & 0.6748 & 0.6733 & 0.2694 & 0.5766 \\
    few-shot       & 0.5584 & 0.5466 & 0.4523 & 0.3731 & 0.7901 & 0.7870 & 0.2357 & 0.6494 & 0.7045 & 0.6999 & 0.2649 & 0.5772 \\
    w/ SFT         & 0.8438 & 0.8160 & 0.2164 & 0.6113 & 0.8263 & 0.8262 & 0.2458 & 0.7107 & 0.8171 & 0.7674 & 0.1957 & 0.6270 \\
    w/ SFT$^*$     & 0.8404 & 0.8031 & \underline{0.1575} & 0.6625 & 0.8243 & 0.8241 & \underline{0.1558} & \underline{0.7355} & 0.8044 & 0.7892 & 0.1727 & 0.6712 \\
    w/ GRPO        & \underline{0.8459} & \underline{0.8179} & 0.2180 & 0.6285 & \underline{0.8323} & \underline{0.8323} & 0.2474 & 0.6851 & \underline{0.8309} & \underline{0.8086} & 0.2022 & 0.6077 \\
    w/ GRPO$^*$    & 0.8408 & 0.8145 & 0.1508 & \underline{0.6704} & 0.8314 & 0.8313 & 0.1622 & 0.7136 & 0.8188 & 0.7992 & \underline{0.1681} & \underline{0.7126} \\
    DPUA           & \textbf{0.8557} & \textbf{0.8218} & \textbf{0.1461} & \textbf{0.7047} & \textbf{0.8378} & \textbf{0.8378} & \textbf{0.1466} & \textbf{0.7697} & \textbf{0.8355} & \textbf{0.8168} & \textbf{0.1598} & \textbf{0.7492} \\
    \midrule
    \multicolumn{13}{c}{LLaMa-3.1-Instruct (8B)} \\
    \midrule
    zero-shot      & 0.5283 & 0.5223 & 0.3936 & 0.1985 & 0.7568 & 0.7533 & 0.2210 & 0.6002 & 0.6830 & 0.6753 & 0.3162 & 0.5034 \\
    few-shot       & 0.5561 & 0.5608 & 0.4458 & 0.2514 & 0.7719 & 0.7709 & 0.2326 & 0.6040 & 0.6899 & 0.6855 & 0.2746 & 0.5549 \\
    w/ SFT         & \underline{0.8512} & 0.7992 & 0.3134 & 0.4720 & 0.8362 & 0.8362 & 0.1878 & 0.7265 & 0.8178 & 0.7978 & 0.2397 & 0.6174 \\
    w/ SFT$^*$     & 0.8333 & 0.7641 & \underline{0.1339} & \underline{0.7258} & 0.8170 & 0.8169 & \underline{0.1360} & \underline{0.7671} & 0.8086 & 0.7729 & 0.1934 & 0.6643 \\
    w/ GRPO        & \underline{0.8512} & \underline{0.8027} & 0.2988 & 0.4460 & \underline{0.8459} & \underline{0.8459} & 0.2110 & 0.6903 & \underline{0.8253} & \underline{0.7997} & 0.2022 & 0.6077 \\
    w/ GRPO$^*$    & 0.8497 & 0.7968 & 0.1371 & 0.7212 & 0.8343 & 0.8343 & 0.1534 & 0.7175 & 0.8171 & 0.7674 & \underline{0.1818} & \underline{0.6698} \\
    DPUA           & \textbf{0.8557} & \textbf{0.8048} & \textbf{0.1259} & \textbf{0.7610} & \textbf{0.8516} & \textbf{0.8514} & \textbf{0.1303} & \textbf{0.7854} & \textbf{0.8332} & \textbf{0.8145} & \textbf{0.1686} & \textbf{0.7215} \\
    \bottomrule
  \end{tabular}
  \label{overall_performance}
\end{table*}
\subsection{Main Result}

Table~\ref{overall_performance} presents the performance of our proposed DPUA framework and the compared baselines on three subjectivity analysis datasets: CSC, MD, and GOE. 
Based on the results, we draw the following conclusions.

\paragraph{Backbone performance under zero-shot settings}
The two backbone models show clear task-dependent differences in the zero-shot setting. On GOE, a sentiment analysis dataset, both models achieve Acc./F1 scores around 0.76, whereas on CSC, a sarcasm detection dataset, their scores drop to around 0.50. This reflects the greater difficulty of sarcasm detection, which requires contextual reasoning over pragmatic incongruity. MD lies between GOE and CSC, as offensiveness detection involves both explicit toxic markers and context-dependent expressions. Notably, uncertainty alignment does not fully follow label prediction performance. For example, although LLaMa-3.1 achieves higher Acc./F1 than Qwen-3 on CSC, its Coef. is much lower, suggesting that better label prediction does not necessarily imply better alignment between model uncertainty and human disagreement. This highlights the distinction between task performance and uncertainty alignment in subjectivity analysis.

\paragraph{Effect of few-shot prompting settings}
Few-shot prompting generally improves task performance, yielding relative gains of 6.41\%/5.96\% in average Acc./F1 for Qwen-3 and 2.53\%/3.40\% for LLaMa-3.1. This suggests that in-context demonstrations help the models better infer task-specific decision boundaries. However, such gains do not transfer consistently to uncertainty alignment. Both backbones show increased MAE under few-shot prompting, indicating less reliable confidence expression. These results suggest that prompting examples mainly benefit label prediction, but are insufficient for guiding models to express uncertainty in accordance with human disagreement.

\paragraph{Effect of conventional training strategies}
Task-specific SFT substantially improves task performance over few-shot prompting, with average Acc./F1 gains of 21.15\%/18.50\% for Qwen-3 and 24.15\%/20.62\% for LLaMa-3.1. It also brings moderate improvements in uncertainty alignment, reducing average MAE by 30.96\% and 22.26\% for the two backbones, respectively. Nevertheless, these improvements are largely indirect effects of supervised learning rather than explicit optimization of uncertainty-aware reasoning or confidence expression. In comparison, vanilla GRPO provides only marginal and inconsistent benefits over SFT. While it slightly improves Acc. and F1, its effects on MAE and Coef. are mixed across backbones. This indicates that conventional reward optimization can strengthen label prediction, but without disagreement-aware rewards, it does not reliably align model uncertainty with human disagreement.

\paragraph{Effect of uncertainty-aware modeling and DPUA.}
Introducing uncertainty-aware modeling leads to clear improvements in uncertainty alignment.
Compared with standard SFT, SFT$^*$ reduces average MAE by 26.13\% for Qwen-3 and 37.47\% for LLaMA-3.1, while improving Coef. by 6.17\% and 18.80\%, respectively.
Similarly, GRPO$^*$ consistently outperforms vanilla GRPO on uncertainty alignment, reducing MAE by 27.94\% and 33.67\% and improving Coef. by 9.12\% and 20.90\% for the two backbones.
These results highlight the importance of explicitly modeling uncertainty reflected by human disagreement.
However, uncertainty-aware baselines sometimes sacrifice task performance, suggesting that simply adding uncertainty expression is insufficient to balance label prediction and uncertainty alignment.

In contrast, the full DPUA framework achieves the most favorable overall trade-off across datasets and backbones.
Compared with the strongest non-DPUA baseline, DPUA maintains competitive label prediction performance, with average Acc./F1 gains of 0.79\%/0.72\% for Qwen-3 and 0.72\%/0.91\% for LLaMA-3.1.
More importantly, it consistently improves uncertainty alignment, reducing average MAE by 5.94\% and 8.31\%, and improving Coef. by 6.06\% and 5.13\%, respectively.
These results indicate that DPUA better balances task performance and uncertainty alignment, demonstrating the benefit of jointly modeling disagreement perception and uncertainty alignment.

\begin{figure*}[t]
    \centering
    \includegraphics[width=1\linewidth]{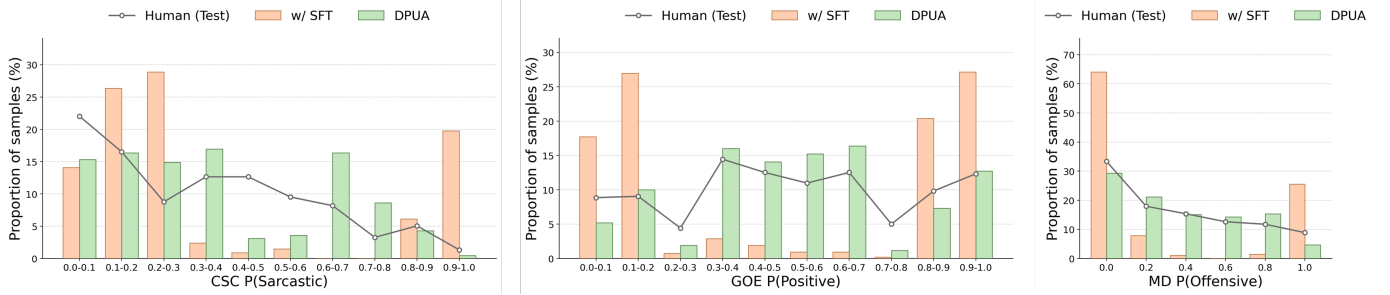}
    \caption{Confidence distribution alignment with human disagreement. For visualization, model confidence scores are converted into class-wise probability distributions following Eq. (8). Compared with SFT, DPUA better aligns the resulting distributions with human disagreement and mitigates overconfidence.}   
\label{distribution}
\end{figure*}

\paragraph{Distribution-level uncertainty alignment}
To further examine whether DPUA aligns model confidence with human disagreement at the distribution level, we convert model confidence scores into class-wise probability distributions following Eq.~(8), and compare them with the human annotation distributions.
As shown in Figure~\ref{distribution}, standard SFT tends to produce more polarized class-wise distributions, assigning many samples to probability regions near 0 or 1.
This indicates that hard-label supervision encourages the model to make overly certain predictions, even when human annotations reveal substantial disagreement.
In contrast, DPUA yields smoother confidence distributions and assigns more samples to intermediate probability regions, which better match the human annotation patterns across the three datasets.
These results suggest that DPUA mitigates overconfidence and better captures the graded uncertainty inherent in subjectivity analysis.

\subsection{Ablation Study}
In this section, we conduct ablation studies to assess the contribution of each key component in DPUA. We consider five variants:
1) \textit{w/o Disagreement Perception (DP)}, which replaces the disagreement-aware training objective in the first phase with standard SFT;
2) \textit{w/o Adaptive Decoupled Learning (DL)}, which replaces adaptive decoupled learning with the general segment-level loss in Eq.~(5);
3) \textit{w/o Uncertainty Alignment (UA)}, which uses the model trained after disagreement perception directly for inference;
4) \textit{w/o Uncertainty-aware Reasoning Reward (URR)}, which removes the uncertainty-aware reasoning reward during the alignment phase;
and 5) \textit{w/o Uncertainty Calibration Reward (UCR)}, which removes the uncertainty calibration reward.
We use Qwen-3 as the backbone and evaluate all variants on CSC and MD, where the backbone shows relatively weaker performance, making these datasets suitable for assessing the contribution of each component.
We report F1 for label prediction and Coef. for uncertainty alignment.

\begin{figure}[t]
    \centering
    \includegraphics[width=0.9\linewidth]{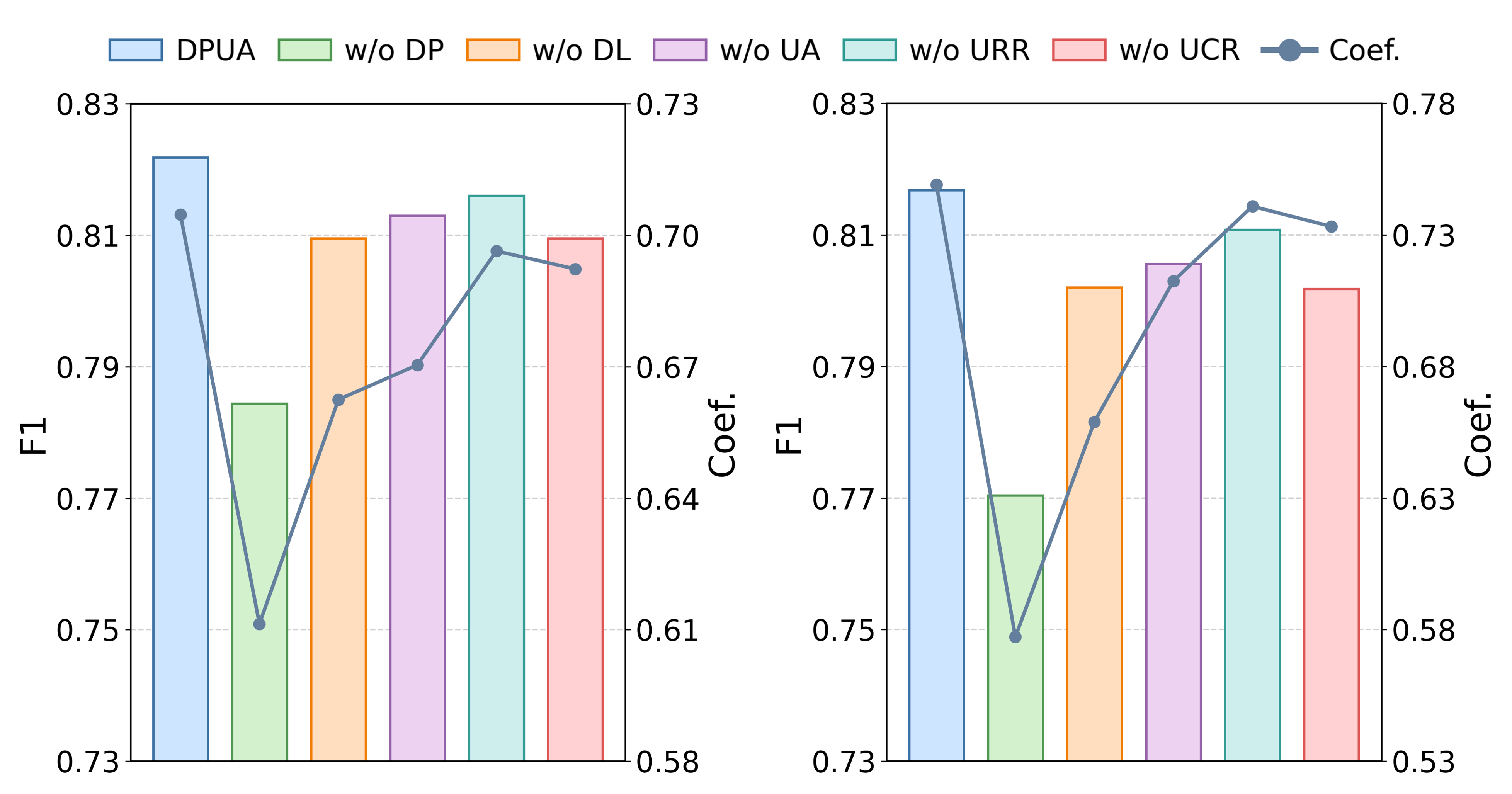}
    \caption{Ablation studies by removing components from DPUA framework.}   
\label{fig:ablation}
\end{figure}

The ablation results in Figure~\ref{fig:ablation} show that each component contributes to DPUA, with different effects on label prediction and uncertainty alignment.
(1) Removing the disagreement perception phase leads to the largest performance drop. Compared with w/o DP, DPUA improves F1 by 4.77\% on CSC and 6.02\% on MD, and improves Coef. by 15.28\% and 19.49\%, respectively. This indicates that explicitly modeling human disagreement in the first phase is crucial for learning transferable uncertainty-aware representations.
(2) Adaptive decoupled learning also plays an important role, especially for uncertainty alignment. Compared with w/o DL, DPUA improves Coef. by 6.37\% on CSC and 13.69\% on MD, while also bringing F1 gains of 1.52\% and 1.85\%. This suggests that decoupling label prediction, rationale learning, and confidence learning helps the model better capture disagreement-related signals.
(3) The uncertainty alignment phase further improves the trade-off between task performance and uncertainty alignment.
Compared with w/o UA, DPUA improves F1 by 1.08\% on CSC and 1.39\% on MD, and improves Coef. by 5.12\% and 5.15\%, respectively.
Among the two rewards, removing either URR or UCR leads to performance degradation, indicating that both rewards contribute to the final alignment stage.
Overall, these results verify that DPUA benefits from both disagreement perception in supervised learning and uncertainty alignment in reward optimization.

\subsection{Generalization Evaluation}

We further examine whether the uncertainty-aware capability learned by DPUA can transfer to out-of-distribution tasks.
Specifically, we directly evaluate models trained on one subjectivity analysis dataset on another dataset, without any additional fine-tuning.
Following the ablation setting, we use CSC and MD as target datasets, where the backbone exhibits weaker uncertainty alignment.
The results are shown in Table~\ref{tab:generalization}.

As shown in Table~\ref{tab:generalization}, DPUA achieves the best or competitive results in most transfer settings. In transfers from GOE to CSC and MD, DPUA consistently improves both label prediction and uncertainty alignment over the zero-shot backbone and SFT-based variants, suggesting that the learned uncertainty-aware capability is not confined to the source task. The gain is particularly clear on GOE $\rightarrow$ CSC, where DPUA improves F1 by 14.68\% over the backbone while also achieving the best Coef. In transfers between CSC and MD, DPUA remains strong under larger task shifts: it obtains the best Acc. and Coef. on MD $\rightarrow$ CSC, and the best Acc., F1, and MAE on CSC $\rightarrow$ MD. This is notable because sarcasm and offensiveness involve different linguistic cues and disagreement sources, requiring the model to generalize beyond surface-level label patterns. Although SFT-based variants perform better on a few individual metrics, they often improve prediction or calibration unevenly. In contrast, DPUA provides a more balanced trade-off between label prediction and uncertainty alignment, suggesting that DPUA captures transferable uncertainty-aware patterns beyond task-specific label distributions.

\begin{table}[t]
\centering
\setlength{\tabcolsep}{9pt}
\caption{Cross-task transfer results on CSC and MD. Qwen-3 denotes the zero-shot backbone evaluated on the target dataset. SFT* denotes the uncertainty-aware SFT variant.}
\label{tab:generalization}
\begin{tabular}{lcccc}
\toprule
Method & Acc. & F1 & MAE & Coef. \\
\midrule
\multicolumn{5}{c}{\textbf{CSC}} \\
\midrule
Qwen-3 & 0.4851 & 0.4850 & 0.4231 & 0.3439 \\
\midrule
\multicolumn{5}{c}{GOE $\rightarrow$ CSC} \\
\cmidrule(lr){1-5}
w/ SFT  & 0.5061 & 0.5166 & 0.3953 & 0.3080 \\
w/ SFT* & 0.5506 & 0.5438 & 0.3640 & 0.3304 \\
DPUA    & \textbf{0.5625} & \textbf{0.5562} & \textbf{0.3011} & \textbf{0.3564} \\
\midrule
\multicolumn{5}{c}{MD $\rightarrow$ CSC} \\
\cmidrule(lr){1-5}
w/ SFT  & 0.7009 & \textbf{0.6438} & 0.2972 & 0.3076 \\
w/ SFT* & 0.7455 & 0.5732 & \textbf{0.2429} & 0.3045 \\
DPUA    & \textbf{0.7485} & 0.5982 & 0.2646 & \textbf{0.3478} \\
\midrule
\multicolumn{5}{c}{\textbf{MD}} \\
\midrule
Qwen-3 & 0.6748 & 0.6733 & 0.2694 & 0.5766 \\
\midrule
\multicolumn{5}{c}{GOE $\rightarrow$ MD} \\
\cmidrule(lr){1-5}
w/ SFT  & 0.6796 & 0.6774 & 0.2900 & 0.5537 \\
w/ SFT* & 0.7017 & 0.6969 & 0.2490 & 0.5930 \\
DPUA    & \textbf{0.7105} & \textbf{0.7050} & \textbf{0.2440} & \textbf{0.5985} \\
\midrule
\multicolumn{5}{c}{CSC $\rightarrow$ MD} \\
\cmidrule(lr){1-5}
w/ SFT  & 0.7370 & 0.7157 & 0.2577 & 0.5213 \\
w/ SFT* & 0.7602 & 0.7204 & 0.2279 & \textbf{0.5399} \\
DPUA    & \textbf{0.7641} & \textbf{0.7270} & \textbf{0.2278} & 0.5321 \\
\bottomrule
\end{tabular}
\end{table}


\subsection{Evaluation of Explainability}
We evaluate the quality of rationales generated by DPUA to further assess its explainability, focusing on whether the rationales can support label prediction and reflect disagreement-related cues.
Since reference-based metrics such as ROUGE and BERT-Score are unsuitable when gold rationales are non-unique, we follow \cite{DBLP:conf/ijcai/WangHACL23} and evaluate rationales along four dimensions:
1) \textit{Informativeness}: whether the rationale provides useful information beyond the prediction, such as relevant context or background;
2) \textit{Readability}: whether the rationale is fluent, grammatical, and easy to understand;
3) \textit{Soundness}: whether the rationale is logically valid and well-supported by the input;
4) \textit{Uncertainty-awareness}: whether the rationale appropriately reflects ambiguity, multiple plausible interpretations, or potential human disagreement.
Each dimension is rated on a 5-point Likert scale, where 1 and 5 indicate the lowest and highest quality, respectively.
We evaluate Qwen-3 rationales on 300 randomly selected test samples from CSC, GOE, and MD, comparing the original backbone, the model with the disagreement perception phase (\textit{w/ DP}), and the full DPUA framework.
For reliability, we conduct both automatic and human evaluation.
GPT-4o is used as the automatic evaluator, while two annotators independently score each rationale for human evaluation, with averaged scores reported.
The results are shown in Table~\ref{evaluation}.

Based on the results, we observe that: 
1) Across most evaluation dimensions, both \textit{w/ DP} and DPUA improve the rationale quality over the original backbone, and the full DPUA framework achieves the best overall performance. This indicates that disagreement-aware training contributes to more reliable explanations for subjectivity analysis.
2) The disagreement perception phase already brings clear gains in \textit{Informativeness} and \textit{Soundness}, suggesting that learning from disagreement signals helps the model identify more relevant evidence and construct more valid rationales.
3) After introducing the full DPUA framework, the improvement becomes more evident, especially in \textit{Uncertainty-awareness}. The average score increases from 2.98 for Qwen to 3.95 for DPUA, corresponding to a relative improvement of 31.2\%. This suggests that DPUA not only improves explanation quality, but also enables the model to better express ambiguity and potential human disagreement.
4) By contrast, the improvement in \textit{Readability} is relatively modest. This is reasonable because the backbone LLM already has strong language generation ability, and therefore its rationales are generally fluent and easy to read even before disagreement-aware optimization.

\begin{figure*}
    \centering
    \includegraphics[width=1.0\linewidth]{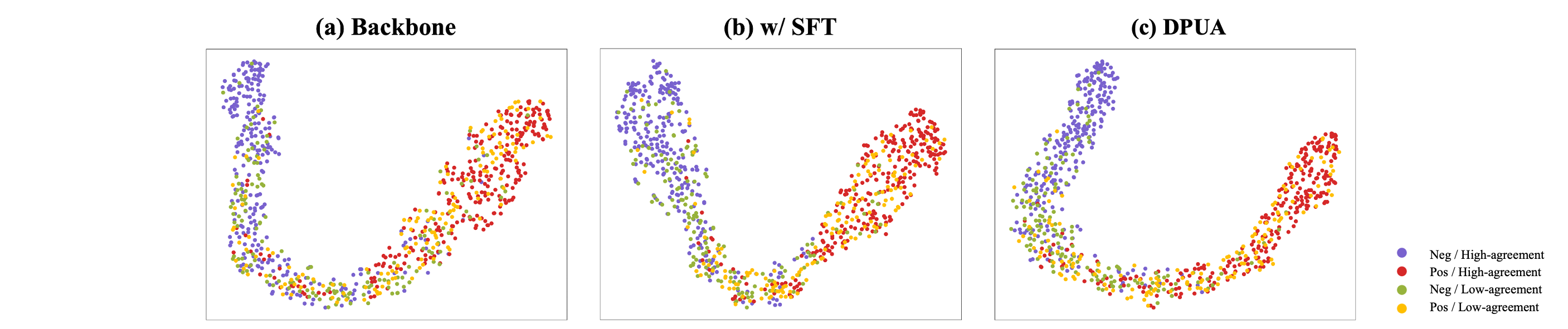}
    \caption{t-SNE visualization of hidden-layer representations on the CSC test set. Compared with the Backbone and w/ SFT settings, DPUA preserves class separability while capturing uncertainty associated with human disagreement near decision boundaries.}
\label{fig:tsne}
\end{figure*}

\begin{table}
  \centering
  \caption{Evaluation of the rationale quality, including Informativeness (Info.), Readability (Read.), Soundness (Sound.), and Uncertainty-awareness (Unc.).}
  \begin{tabular}{
    >{\centering\arraybackslash}m{1.25cm}
    >{\centering\arraybackslash}m{0.75cm}
    >{\centering\arraybackslash}m{0.75cm}
    >{\centering\arraybackslash}m{0.75cm}
    >{\centering\arraybackslash}m{0.75cm}
    >{\centering\arraybackslash}m{0.75cm}
    >{\centering\arraybackslash}m{0.75cm}
  }
    \toprule \multicolumn{1}{c}{Evaluator} & \multicolumn{3}{c}{GPT-4o} & \multicolumn{3}{c}{Human} \\
    \cmidrule(lr){2-4}
    \cmidrule(lr){5-7}
    Rationales & Qwen  & w/ DP & DPUA  & Qwen  & w/ DP & DPUA \\
    \midrule
    Info. & 3.45  & 3.95  & 3.98  & 3.39  & 4.08  & 4.15 \\
    Read. & 4.28  & 4.30  & 4.30  & 4.82  & 4.84  & 4.82 \\
    Sound. & 3.04  & 3.75  & 3.93  & 2.78  & 3.52  & 3.73 \\
    Unc. & 3.10  & 3.52  & 3.91  & 2.85  & 3.64  & 3.99 \\
    \bottomrule
    \end{tabular}%
  \label{evaluation}%
\end{table}%

\subsection{Representation Analysis}
To further investigate the mechanism of DPUA, we extract the hidden-layer representations of Qwen-3 on the CSC test set and project them into a two-dimensional space using t-SNE. 
The same procedure is applied to the Backbone and w/ SFT settings, allowing us to compare the representation structures learned by different methods and examine their effects on class separability and uncertainty modeling. 
For clarity, we further partition the samples according to their human agreement scores, where samples with scores greater than 0.75 are treated as high-agreement samples, while those with scores no greater than 0.75 are treated as low-agreement samples. 
The results are shown in Figure~\ref{fig:tsne}.

Compared with the backbone, SFT improves the model's ability to discriminate task labels, making the separation between positive and negative samples more pronounced in the representation space.
However, since SFT mainly relies on hard-label supervision, it tends to pull samples toward the interior of class manifolds and compress the natural transition structure around decision boundaries.
This representation pattern may induce overconfident predictions on low-agreement samples, where human judgments are inherently less consistent.

In contrast, DPUA preserves class separability while maintaining smoother and less rigid boundary structures.
High-agreement samples are mainly located in the interior regions of each class manifold, whereas low-agreement samples are more often distributed near the boundaries between class manifolds.
This visualization provides qualitative evidence that DPUA not only learns task-discriminative representations, but also captures uncertainty associated with human disagreement.
Such a structure may partially explain its cross-task transferability: instead of merely fitting dataset-specific labels, DPUA learns a more general representation of subjective ambiguity, where high-agreement and low-agreement samples occupy relatively distinct regions of the decision space.
These observations partially explain why DPUA mitigates overconfidence on boundary samples and supports more transferable uncertainty-aware subjectivity analysis.

\subsection{Case Study}

To examine how DPUA aligns LLM uncertainty with human disagreement, we conduct case studies on three instances, each randomly sampled from a different dataset.
We compare DPUA with the original Qwen-3 backbone in terms of prediction, rationale, and confidence score.
Based on Table~\ref{case_study}, we draw the following observations.

First, DPUA preserves downstream task performance.
Across all three cases, DPUA produces correct predictions consistent with the ground-truth labels, matching or improving upon the original Qwen-3 backbone.
For example, in Case 1, DPUA correctly identifies the offensiveness of the expression \textit{``Soy version''}, whereas the backbone fails to recognize its underlying mocking connotation.
These results suggest that introducing uncertainty-aware learning does not compromise the model's ability to make accurate decisions.

Second, DPUA aligns its confidence more closely with human agreement.
Compared with Qwen-3, DPUA assigns lower confidence to more ambiguous cases and explicitly reflects alternative interpretations in its rationales.
For instance, it recognizes that \textit{``HELLOOOOO STEVE!''} in Case 2 may be interpreted as either sarcasm or playful excitement, and that \textit{``Imagine being upset over who someone loves''} in Case 3 may convey both negative criticism and implicit support for love acceptance.
Overall, these examples show that DPUA maintains reliable predictions while expressing uncertainty in a manner better aligned with human disagreement.

\begin{table*}
\centering
\setlength{\tabcolsep}{4pt}
\renewcommand{\arraystretch}{1.08}
\caption{Illustration of the case study.}
\label{case_study}

\begin{tabular}{p{0.075\linewidth} p{0.85\linewidth}}
\toprule
\multicolumn{2}{l}{\textbf{Case 1: Offensiveness Detection (MD)}} \\
\midrule
\textbf{Sample} &
\textit{``Why does the mayor look like a Soy version of Justin Trudeau.''} \quad \textbf{Label:} Offensive \quad \textbf{Agreement Score:} 0.60 \\

\textbf{Qwen-3} &
\textbf{Prediction:} Normal  \quad  \textbf{Confidence:} 0.95 \newline
\textbf{Rationale:} The text does not contain explicit markers of offense. The phrase "Soy version" potentially indicates a youthful appearance, which does not inherently insult or debase any individual or group.\\

\textbf{DPUA} &
\textbf{Prediction:} Offensive \quad  \textbf{Confidence:} 0.75 \newline
\textbf{Rationale:} The text is offensive because “\textit{Soy}” is often used pejoratively in political discourse to mock masculinity. However, disagreement may arise because some readers may interpret it as black humor.\\

\midrule
\multicolumn{2}{l}{\textbf{Case 2: Sarcasm Detection (CSC)}} \\
\midrule
\textbf{Sample} &
\textbf{Context:} \textit{``You walk into the room and Steve is there and Steve says "hi!"''}  \textbf{Response:} \textit{``HELLOOOOO STEVE!''} \newline \textbf{Label:} Sarcastic \quad \textbf{Agreement Score:} 0.60 \\

\textbf{Qwen-3} &
\textbf{Prediction:} Sarcastic \quad \textbf{Confidence:} 0.88 \newline
\textbf{Rationale:} The response is sarcastic through excessive capitalization and elongated letters \textit{”O“}, creating an exaggerated and potentially mocking tone.\\

\textbf{DPUA} &
\textbf{Prediction:} Sarcastic \quad \textbf{Confidence:} 0.64 \newline
\textbf{Rationale:} This sample is sarcastic because the exaggerated response contrasts with Steve’s casual \textit{“hi!”}, suggesting mockery or over-performance. However, some may interpret it simply as playful excitement.\\

\midrule
\multicolumn{2}{l}{\textbf{Case 3: Sentiment Analysis (GOE)}} \\
\midrule
\textbf{Sample} &
\textit{``Imagine being upset over who someone loves.''} \quad \textbf{Label:} Negative \quad \textbf{Agreement Score:} 0.60 \\

\textbf{Qwen-3} &
\textbf{Prediction:} Negative \quad  \textbf{Confidence:} 0.85 \newline
\textbf{Rationale:} This text is negative because it criticizes people who are upset over someone’s choice of whom to love. The word “upset” signals a negative emotional state.\\

\textbf{DPUA} &
\textbf{Prediction:} Negative \quad  \textbf{Confidence:} 0.66 \newline
\textbf{Rationale:} This text is negative because it expresses mild criticism toward those who are upset over someone’s choice of whom to love. However, some may interpret it as positive due to its implicit support for love acceptance.\\

\bottomrule
\end{tabular}
\end{table*}

\section{Conclusion}

In this paper, we study how to mitigate LLM overconfidence in subjectivity analysis by enabling models to make accurate label predictions while expressing uncertainty aligned with human disagreement. To this end, we propose Disagreement Perception and Uncertainty Alignment (DPUA), a two-phase framework for uncertainty-aware modeling.
DPUA first learns disagreement signals and then further aligns reasoning and uncertainty expression with the underlying human disagreement distribution. 
Experimental results across multiple subjectivity analysis datasets show that DPUA preserves core task performance, improves the alignment between model uncertainty and human disagreement, and reduces overconfidence on boundary samples. 
Additional evaluations under out-of-distribution settings further suggest that the ability to align uncertainty with human disagreement is transferable and supports better generalization. 
Overall, our findings underscore the value of modeling human disagreement as an intrinsic source of uncertainty and highlight its importance for developing more trustworthy and human-aligned LLMs in subjective decision-making scenarios.

\section{Acknowledgment}
This research was supported in part by the National Natural Science Foundation of China (Nos. 625B2033, 62576073, and 62376051), the Ministry of Education Humanities and Social Science Project of China (No. 25YJCZH308).
We sincerely thank the reviewers for their valuable comments and constructive suggestions, which have helped improve the quality of this paper.

\bibliographystyle{IEEEtran}
\bibliography{LLM, uncertain, base_hate}

@inproceedings{DBLP:conf/nips/KojimaGRMI22,
  author       = {Takeshi Kojima and
                  Shixiang Shane Gu and
                  Machel Reid and
                  Yutaka Matsuo and
                  Yusuke Iwasawa},
  editor       = {Sanmi Koyejo and
                  S. Mohamed and
                  A. Agarwal and
                  Danielle Belgrave and
                  K. Cho and
                  A. Oh},
  title        = {Large Language Models are Zero-Shot Reasoners},
  booktitle    = {Advances in Neural Information Processing Systems 35: Annual Conference
                  on Neural Information Processing Systems 2022, NeurIPS 2022, New Orleans,
                  LA, USA, November 28 - December 9, 2022},
  year         = {2022},
  url          = {http://papers.nips.cc/paper\_files/paper/2022/hash/8bb0d291acd4acf06ef112099c16f326-Abstract-Conference.html},
  bibsource    = {dblp computer science bibliography, https://dblp.org}
}

@article{DBLP:journals/jmlr/ChowdheryNDBMRBCSGSSTMRBTSPRDHPBAI23,
  author       = {Aakanksha Chowdhery and
                  Sharan Narang and
                  Jacob Devlin and
                  Maarten Bosma and
                  Gaurav Mishra and
                  Adam Roberts and
                  Paul Barham and
                  Hyung Won Chung and
                  Charles Sutton and
                  Sebastian Gehrmann and
                  Parker Schuh and
                  Kensen Shi and
                  Sasha Tsvyashchenko and
                  Joshua Maynez and
                  Abhishek Rao and
                  Parker Barnes and
                  Yi Tay and
                  Noam Shazeer and
                  Vinodkumar Prabhakaran and
                  Emily Reif and
                  Nan Du and
                  Ben Hutchinson and
                  Reiner Pope and
                  James Bradbury and
                  Jacob Austin and
                  Michael Isard and
                  Guy Gur{-}Ari and
                  Pengcheng Yin and
                  Toju Duke and
                  Anselm Levskaya and
                  Sanjay Ghemawat and
                  Sunipa Dev and
                  Henryk Michalewski and
                  Xavier Garcia and
                  Vedant Misra and
                  Kevin Robinson and
                  Liam Fedus and
                  Denny Zhou and
                  Daphne Ippolito and
                  David Luan and
                  Hyeontaek Lim and
                  Barret Zoph and
                  Alexander Spiridonov and
                  Ryan Sepassi and
                  David Dohan and
                  Shivani Agrawal and
                  Mark Omernick and
                  Andrew M. Dai and
                  Thanumalayan Sankaranarayana Pillai and
                  Marie Pellat and
                  Aitor Lewkowycz and
                  Erica Moreira and
                  Rewon Child and
                  Oleksandr Polozov and
                  Katherine Lee and
                  Zongwei Zhou and
                  Xuezhi Wang and
                  Brennan Saeta and
                  Mark Diaz and
                  Orhan Firat and
                  Michele Catasta and
                  Jason Wei and
                  Kathy Meier{-}Hellstern and
                  Douglas Eck and
                  Jeff Dean and
                  Slav Petrov and
                  Noah Fiedel},
  title        = {PaLM: Scaling Language Modeling with Pathways},
  journal      = {J. Mach. Learn. Res.},
  volume       = {24},
  pages        = {240:1--240:113},
  year         = {2023},
  url          = {http://jmlr.org/papers/v24/22-1144.html},
  bibsource    = {dblp computer science bibliography, https://dblp.org}
}

@article{DBLP:journals/corr/abs-2303-08774,
  author       = {OpenAI},
  title        = {{GPT-4} Technical Report},
  journal      = {CoRR},
  volume       = {abs/2303.08774},
  year         = {2023},
  url          = {https://doi.org/10.48550/arXiv.2303.08774},
  doi          = {10.48550/ARXIV.2303.08774},
  eprinttype    = {arXiv},
  eprint       = {2303.08774},
  bibsource    = {dblp computer science bibliography, https://dblp.org}
}

@article{DBLP:journals/corr/abs-2502-04194,
  author       = {Dylan Zhang and
                  Qirun Dai and
                  Hao Peng},
  title        = {The Best Instruction-Tuning Data are Those That Fit},
  journal      = {CoRR},
  volume       = {abs/2502.04194},
  year         = {2025},
  url          = {https://doi.org/10.48550/arXiv.2502.04194},
  doi          = {10.48550/ARXIV.2502.04194},
  eprinttype   = {arXiv},
  eprint       = {2502.04194},
  bibsource    = {dblp computer science bibliography, https://dblp.org}
}

@inproceedings{DBLP:conf/iclr/Zhou0NLWWHWH25,
  author       = {Zihao Zhou and
                  Shudong Liu and
                  Maizhen Ning and
                  Wei Liu and
                  Jindong Wang and
                  Derek F. Wong and
                  Xiaowei Huang and
                  Qiufeng Wang and
                  Kaizhu Huang},
  title        = {Is Your Model Really {A} Good Math Reasoner? Evaluating Mathematical
                  Reasoning with Checklist},
  booktitle    = {The Thirteenth International Conference on Learning Representations,
                  {ICLR} 2025, Singapore, April 24-28, 2025},
  publisher    = {OpenReview.net},
  year         = {2025},
  url          = {https://openreview.net/forum?id=nDvgHIBRxQ},
  bibsource    = {dblp computer science bibliography, https://dblp.org}
}

@inproceedings{DBLP:conf/iclr/HuCLGWYG24,
  author       = {Xuming Hu and
                  Junzhe Chen and
                  Xiaochuan Li and
                  Yufei Guo and
                  Lijie Wen and
                  Philip S. Yu and
                  Zhijiang Guo},
  title        = {Towards Understanding Factual Knowledge of Large Language Models},
  booktitle    = {The Twelfth International Conference on Learning Representations,
                  {ICLR} 2024, Vienna, Austria, May 7-11, 2024},
  publisher    = {OpenReview.net},
  year         = {2024},
  url          = {https://openreview.net/forum?id=9OevMUdods},
  bibsource    = {dblp computer science bibliography, https://dblp.org}
}

@article{DBLP:journals/corr/abs-2402-03300,
  author       = {Zhihong Shao and
                  Peiyi Wang and
                  Qihao Zhu and
                  Runxin Xu and
                  Junxiao Song and
                  Mingchuan Zhang and
                  Y. K. Li and
                  Y. Wu and
                  Daya Guo},
  title        = {DeepSeekMath: Pushing the Limits of Mathematical Reasoning in Open
                  Language Models},
  journal      = {CoRR},
  volume       = {abs/2402.03300},
  year         = {2024},
  url          = {https://doi.org/10.48550/arXiv.2402.03300},
  doi          = {10.48550/ARXIV.2402.03300},
  eprinttype   = {arXiv},
  eprint       = {2402.03300},
  bibsource    = {dblp computer science bibliography, https://dblp.org}
}

@inproceedings{DBLP:conf/icwsm/DavidsonWMW17,
  author       = {Thomas Davidson and
                  Dana Warmsley and
                  Michael W. Macy and
                  Ingmar Weber},
  title        = {Automated Hate Speech Detection and the Problem of Offensive Language},
  booktitle    = {Proceedings of the Eleventh International Conference on Web and Social
                  Media, {ICWSM} 2017, Montr{\'{e}}al, Qu{\'{e}}bec, Canada,
                  May 15-18, 2017},
  pages        = {512--515},
  publisher    = {{AAAI} Press},
  year         = {2017},
  url          = {https://aaai.org/ocs/index.php/ICWSM/ICWSM17/paper/view/15665},
  bibsource    = {dblp computer science bibliography, https://dblp.org}
}

@inproceedings{DBLP:conf/aaai/MathewSYBG021,
  author    = {Binny Mathew and
               Punyajoy Saha and
               Seid Muhie Yimam and
               Chris Biemann and
               Pawan Goyal and
               Animesh Mukherjee},
  title     = {HateXplain: {A} Benchmark Dataset for Explainable Hate Speech Detection},
  booktitle = {Thirty-Fifth {AAAI} Conference on Artificial Intelligence, {AAAI}
               2021, Thirty-Third Conference on Innovative Applications of Artificial
               Intelligence, {IAAI} 2021, The Eleventh Symposium on Educational Advances
               in Artificial Intelligence, {EAAI} 2021, Virtual Event, February 2-9,
               2021},
  pages     = {14867--14875},
  publisher = {{AAAI} Press},
  year      = {2021},
  url       = {https://ojs.aaai.org/index.php/AAAI/article/view/17745},
  bibsource = {dblp computer science bibliography, https://dblp.org}
}

@inproceedings{DBLP:conf/ijcai/WangHACL23,
  author       = {Han Wang and
                  Ming Shan Hee and
                  Md. Rabiul Awal and
                  Kenny Tsu Wei Choo and
                  Roy Ka{-}Wei Lee},
  title        = {Evaluating {GPT-3} Generated Explanations for Hateful Content Moderation},
  booktitle    = {Proceedings of the Thirty-Second International Joint Conference on
                  Artificial Intelligence, {IJCAI} 2023, 19th-25th August 2023, Macao,
                  SAR, China},
  pages        = {6255--6263},
  publisher    = {ijcai.org},
  year         = {2023},
}

@inproceedings{DBLP:conf/www/AroyoDTRR19,
  author       = {Lora Aroyo and
                  Lucas Dixon and
                  Nithum Thain and
                  Olivia Redfield and
                  Rachel Rosen},
  editor       = {Sihem Amer{-}Yahia and
                  Mohammad Mahdian and
                  Ashish Goel and
                  Geert{-}Jan Houben and
                  Kristina Lerman and
                  Julian J. McAuley and
                  Ricardo Baeza{-}Yates and
                  Leila Zia},
  title        = {Crowdsourcing Subjective Tasks: The Case Study of Understanding Toxicity
                  in Online Discussions},
  booktitle    = {Companion of The 2019 World Wide Web Conference, {WWW} 2019, San Francisco,
                  CA, USA, May 13-17, 2019},
  pages        = {1100--1105},
  publisher    = {{ACM}},
  year         = {2019},
  url          = {https://doi.org/10.1145/3308560.3317083},
  doi          = {10.1145/3308560.3317083},
  bibsource    = {dblp computer science bibliography, https://dblp.org}
}

@inproceedings{basile-etal-2021-need,
    title = "We Need to Consider Disagreement in Evaluation",
    author = "Basile, Valerio  and
      Fell, Michael  and
      Fornaciari, Tommaso  and
      Hovy, Dirk  and
      Paun, Silviu  and
      Plank, Barbara  and
      Poesio, Massimo  and
      Uma, Alexandra",
    editor = "Church, Kenneth  and
      Liberman, Mark  and
      Kordoni, Valia",
    booktitle = "Proceedings of the 1st Workshop on Benchmarking: Past, Present and Future",
    month = aug,
    year = "2021",
    address = "Online",
    publisher = "Association for Computational Linguistics",
    url = "https://aclanthology.org/2021.bppf-1.3/",
    doi = "10.18653/v1/2021.bppf-1.3",
    pages = "15--21",
}

@inproceedings{DBLP:conf/aiia/Basile20,
  author       = {Valerio Basile},
  editor       = {Giuseppe Vizzari and
                  Matteo Palmonari and
                  Andrea Orlandini},
  title        = {It's the End of the Gold Standard as we Know it. On the Impact of
               Pre-aggregation on the Evaluation of Highly Subjective Tasks},
  booktitle    = {Proceedings of the AIxIA 2020 Discussion Papers Workshop co-located
                  with the the 19th International Conference of the Italian Association
                  for Artificial Intelligence (AIxIA2020), Anywhere, November 27th,
                  2020},
  series       = {{CEUR} Workshop Proceedings},
  volume       = {2776},
  pages        = {31--40},
  publisher    = {CEUR-WS.org},
  year         = {2020},
  url          = {https://ceur-ws.org/Vol-2776/paper-4.pdf},
  bibsource    = {dblp computer science bibliography, https://dblp.org}
}

@inproceedings{uma-etal-2021-semeval,
    title = "{S}em{E}val-2021 Task 12: Learning with Disagreements",
    author = "Uma, Alexandra  and
      Fornaciari, Tommaso  and
      Dumitrache, Anca  and
      Miller, Tristan  and
      Chamberlain, Jon  and
      Plank, Barbara  and
      Simpson, Edwin  and
      Poesio, Massimo",
    booktitle = "Proceedings of the 15th International Workshop on Semantic Evaluation (SemEval-2021)",
    month = aug,
    year = "2021",
    address = "Online",
    publisher = "Association for Computational Linguistics",
    url = "https://aclanthology.org/2021.semeval-1.41/",
    doi = "10.18653/v1/2021.semeval-1.41",
    pages = "338--347",
}

@inproceedings{DBLP:conf/semeval/LeonardelliAABF23,
  author       = {Elisa Leonardelli and
                  Gavin Abercrombie and
                  Dina Almanea and
                  Valerio Basile and
                  Tommaso Fornaciari and
                  Barbara Plank and
                  Verena Rieser and
                  Alexandra Uma and
                  Massimo Poesio},
  editor       = {Atul Kr. Ojha and
                  A. Seza Dogru{\"{o}}z and
                  Giovanni Da San Martino and
                  Harish Tayyar Madabushi and
                  Ritesh Kumar and
                  Elisa Sartori},
  title        = {SemEval-2023 Task 11: Learning with Disagreements (LeWiDi)},
  booktitle    = {Proceedings of the The 17th International Workshop on Semantic Evaluation,
                  SemEval@ACL 2023, Toronto, Canada, 13-14 July 2023},
  pages        = {2304--2318},
  publisher    = {Association for Computational Linguistics},
  year         = {2023},
  url          = {https://doi.org/10.18653/v1/2023.semeval-1.314},
  doi          = {10.18653/V1/2023.SEMEVAL-1.314},
  bibsource    = {dblp computer science bibliography, https://dblp.org}
}

@inproceedings{DBLP:conf/eacl/SandriLTJ23,
  author       = {Marta Sandri and
                  Elisa Leonardelli and
                  Sara Tonelli and
                  Elisabetta Jezek},
  editor       = {Andreas Vlachos and
                  Isabelle Augenstein},
  title        = {Why Don't You Do It Right? Analysing Annotators' Disagreement in Subjective
                  Tasks},
  booktitle    = {Proceedings of the 17th Conference of the European Chapter of the
                  Association for Computational Linguistics, {EACL} 2023, Dubrovnik,
                  Croatia, May 2-6, 2023},
  pages        = {2420--2433},
  publisher    = {Association for Computational Linguistics},
  year         = {2023},
  url          = {https://doi.org/10.18653/v1/2023.eacl-main.178},
  doi          = {10.18653/V1/2023.EACL-MAIN.178},
  bibsource    = {dblp computer science bibliography, https://dblp.org}
}

@inproceedings{DBLP:conf/emnlp/DengZ0W0M23,
  author       = {Naihao Deng and
                  Xinliang Frederick Zhang and
                  Siyang Liu and
                  Winston Wu and
                  Lu Wang and
                  Rada Mihalcea},
  editor       = {Houda Bouamor and
                  Juan Pino and
                  Kalika Bali},
  title        = {You Are What You Annotate: Towards Better Models through Annotator
                  Representations},
  booktitle    = {Findings of the Association for Computational Linguistics: {EMNLP}
                  2023, Singapore, December 6-10, 2023},
  pages        = {12475--12498},
  publisher    = {Association for Computational Linguistics},
  year         = {2023},
  url          = {https://doi.org/10.18653/v1/2023.findings-emnlp.832},
  doi          = {10.18653/V1/2023.FINDINGS-EMNLP.832},
  bibsource    = {dblp computer science bibliography, https://dblp.org}
}

@inproceedings{DBLP:conf/acl/Chen024,
  author       = {Jiuhai Chen and
                  Jonas Mueller},
  editor       = {Lun{-}Wei Ku and
                  Andre Martins and
                  Vivek Srikumar},
  title        = {Quantifying Uncertainty in Answers from any Language Model and Enhancing
                  their Trustworthiness},
  booktitle    = {Proceedings of the 62nd Annual Meeting of the Association for Computational
                  Linguistics (Volume 1: Long Papers), {ACL} 2024, Bangkok, Thailand,
                  August 11-16, 2024},
  pages        = {5186--5200},
  publisher    = {Association for Computational Linguistics},
  year         = {2024},
  url          = {https://doi.org/10.18653/v1/2024.acl-long.283},
  doi          = {10.18653/V1/2024.ACL-LONG.283},
  bibsource    = {dblp computer science bibliography, https://dblp.org}
}

@inproceedings{DBLP:conf/iclr/0002WSLCNCZ23,
  author       = {Xuezhi Wang and
                  Jason Wei and
                  Dale Schuurmans and
                  Quoc V. Le and
                  Ed H. Chi and
                  Sharan Narang and
                  Aakanksha Chowdhery and
                  Denny Zhou},
  title        = {Self-Consistency Improves Chain of Thought Reasoning in Language Models},
  booktitle    = {The Eleventh International Conference on Learning Representations,
                  {ICLR} 2023, Kigali, Rwanda, May 1-5, 2023},
  publisher    = {OpenReview.net},
  year         = {2023},
  url          = {https://openreview.net/forum?id=1PL1NIMMrw},
  bibsource    = {dblp computer science bibliography, https://dblp.org}
}

@inproceedings{DBLP:conf/aaai/0001SM0ETASC25,
  author       = {Qing Lyu and
                  Kumar Shridhar and
                  Chaitanya Malaviya and
                  Li Zhang and
                  Yanai Elazar and
                  Niket Tandon and
                  Marianna Apidianaki and
                  Mrinmaya Sachan and
                  Chris Callison{-}Burch},
  editor       = {Toby Walsh and
                  Julie Shah and
                  Zico Kolter},
  title        = {Calibrating Large Language Models with Sample Consistency},
  booktitle    = {Thirty-Ninth {AAAI} Conference on Artificial Intelligence, Thirty-Seventh
                  Conference on Innovative Applications of Artificial Intelligence,
                  Fifteenth Symposium on Educational Advances in Artificial Intelligence,
                  {AAAI} 2025, Philadelphia, PA, USA, February 25 - March 4, 2025},
  pages        = {19260--19268},
  publisher    = {{AAAI} Press},
  year         = {2025},
  url          = {https://doi.org/10.1609/aaai.v39i18.34120},
  doi          = {10.1609/AAAI.V39I18.34120},
  bibsource    = {dblp computer science bibliography, https://dblp.org}
}

@article{DBLP:journals/tmlr/LinHE22,
  author       = {Stephanie Lin and
                  Jacob Hilton and
                  Owain Evans},
  title        = {Teaching Models to Express Their Uncertainty in Words},
  journal      = {Trans. Mach. Learn. Res.},
  volume       = {2022},
  year         = {2022},
  url          = {https://openreview.net/forum?id=8s8K2UZGTZ},
  bibsource    = {dblp computer science bibliography, https://dblp.org}
}

@inproceedings{plank-2022-problem,
    title = "The {\textquotedblleft}Problem{\textquotedblright} of Human Label Variation: On Ground Truth in Data, Modeling and Evaluation",
    author = "Plank, Barbara",
    editor = "Goldberg, Yoav  and
      Kozareva, Zornitsa  and
      Zhang, Yue",
    booktitle = "Proceedings of the 2022 Conference on Empirical Methods in Natural Language Processing",
    month = dec,
    year = "2022",
    address = "Abu Dhabi, United Arab Emirates",
    publisher = "Association for Computational Linguistics",
    url = "https://aclanthology.org/2022.emnlp-main.731/",
    doi = "10.18653/v1/2022.emnlp-main.731",
    pages = "10671--10682",
}

@article{DBLP:journals/jair/UmaFHPPP21,
  author       = {Alexandra Uma and
                  Tommaso Fornaciari and
                  Dirk Hovy and
                  Silviu Paun and
                  Barbara Plank and
                  Massimo Poesio},
  title        = {Learning from Disagreement: {A} Survey},
  journal      = {J. Artif. Intell. Res.},
  volume       = {72},
  pages        = {1385--1470},
  year         = {2021},
  url          = {https://doi.org/10.1613/jair.1.12752},
  doi          = {10.1613/JAIR.1.12752},
  bibsource    = {dblp computer science bibliography, https://dblp.org}
}

@article{DBLP:journals/tacl/DavaniDP22,
  author       = {Aida Mostafazadeh Davani and
                  Mark D{\'{\i}}az and
                  Vinodkumar Prabhakaran},
  title        = {Dealing with Disagreements: Looking Beyond the Majority Vote in Subjective
                  Annotations},
  journal      = {Trans. Assoc. Comput. Linguistics},
  volume       = {10},
  pages        = {92--110},
  year         = {2022},
  url          = {https://doi.org/10.1162/tacl\_a\_00449},
  doi          = {10.1162/TACL\_A\_00449},
  bibsource    = {dblp computer science bibliography, https://dblp.org}
}

@inproceedings{weerasooriya-etal-2023-disagreement,
    title = "Disagreement Matters: Preserving Label Diversity by Jointly Modeling Item and Annotator Label Distributions with {D}is{C}o",
    author = "Weerasooriya, Tharindu Cyril  and
      Ororbia, Alexander  and
      Bhensadadia, Raj  and
      KhudaBukhsh, Ashiqur  and
      Homan, Christopher",
    booktitle = "Findings of the Association for Computational Linguistics: ACL 2023",
    month = jul,
    year = "2023",
    address = "Toronto, Canada",
    publisher = "Association for Computational Linguistics",
    url = "https://aclanthology.org/2023.findings-acl.287/",
    doi = "10.18653/v1/2023.findings-acl.287",
    pages = "4679--4695",
}

@inproceedings{DBLP:conf/emnlp/ParappanH25,
  author       = {Mohammed Fayiz Parappan and
                  Ricardo Henao},
  editor       = {Christos Christodoulopoulos and
                  Tanmoy Chakraborty and
                  Carolyn Rose and
                  Violet Peng},
  title        = {Learning Subjective Label Distributions via Sociocultural Descriptors},
  booktitle    = {Proceedings of the 2025 Conference on Empirical Methods in Natural
                  Language Processing, {EMNLP} 2025, Suzhou, China, November 4-9, 2025},
  pages        = {20322--20338},
  publisher    = {Association for Computational Linguistics},
  year         = {2025},
  url          = {https://doi.org/10.18653/v1/2025.emnlp-main.1026},
  doi          = {10.18653/V1/2025.EMNLP-MAIN.1026},
  bibsource    = {dblp computer science bibliography, https://dblp.org}
}

@inproceedings{DBLP:conf/naacl/FornaciariUPPHP21,
  author       = {Tommaso Fornaciari and
                  Alexandra Uma and
                  Silviu Paun and
                  Barbara Plank and
                  Dirk Hovy and
                  Massimo Poesio},
  editor       = {Kristina Toutanova and
                  Anna Rumshisky and
                  Luke Zettlemoyer and
                  Dilek Hakkani{-}T{\"{u}}r and
                  Iz Beltagy and
                  Steven Bethard and
                  Ryan Cotterell and
                  Tanmoy Chakraborty and
                  Yichao Zhou},
  title        = {Beyond Black {\&} White: Leveraging Annotator Disagreement via
                  Soft-Label Multi-Task Learning},
  booktitle    = {Proceedings of the 2021 Conference of the North American Chapter of
                  the Association for Computational Linguistics: Human Language Technologies,
                  {NAACL-HLT} 2021, Online, June 6-11, 2021},
  pages        = {2591--2597},
  publisher    = {Association for Computational Linguistics},
  year         = {2021},
  url          = {https://doi.org/10.18653/v1/2021.naacl-main.204},
  doi          = {10.18653/V1/2021.NAACL-MAIN.204},
  bibsource    = {dblp computer science bibliography, https://dblp.org}
}

@inproceedings{DBLP:conf/acl/BaumlerSD23,
  author       = {Connor Baumler and
                  Anna Sotnikova and
                  Hal Daum{\'{e}} III},
  editor       = {Anna Rogers and
                  Jordan L. Boyd{-}Graber and
                  Naoaki Okazaki},
  title        = {Which Examples Should be Multiply Annotated? Active Learning When
                  Annotators May Disagree},
  booktitle    = {Findings of the Association for Computational Linguistics: {ACL} 2023,
                  Toronto, Canada, July 9-14, 2023},
  series       = {Findings of {ACL}},
  pages        = {10352--10371},
  publisher    = {Association for Computational Linguistics},
  year         = {2023},
  url          = {https://doi.org/10.18653/v1/2023.findings-acl.658},
  doi          = {10.18653/V1/2023.FINDINGS-ACL.658},
  bibsource    = {dblp computer science bibliography, https://dblp.org}
}

@inproceedings{DBLP:conf/acl/LuMWXL00L25,
  author       = {Junyu Lu and
                  Kai Ma and
                  Kaichun Wang and
                  Kelaiti Xiao and
                  Roy Ka{-}Wei Lee and
                  Bo Xu and
                  Liang Yang and
                  Hongfei Lin},
  editor       = {Wanxiang Che and
                  Joyce Nabende and
                  Ekaterina Shutova and
                  Mohammad Taher Pilehvar},
  title        = {Is {LLM} an Overconfident Judge? Unveiling the Capabilities of LLMs
                  in Detecting Offensive Language with Annotation Disagreement},
  booktitle    = {Findings of the Association for Computational Linguistics, {ACL} 2025,
                  Vienna, Austria, July 27 - August 1, 2025},
  series       = {Findings of {ACL}},
  pages        = {5609--5626},
  publisher    = {Association for Computational Linguistics},
  year         = {2025},
  url          = {https://aclanthology.org/2025.findings-acl.293/},
  bibsource    = {dblp computer science bibliography, https://dblp.org}
}

@inproceedings{DBLP:conf/acl/DemszkyMKCNR20,
  author       = {Dorottya Demszky and
                  Dana Movshovitz{-}Attias and
                  Jeongwoo Ko and
                  Alan S. Cowen and
                  Gaurav Nemade and
                  Sujith Ravi},
  editor       = {Dan Jurafsky and
                  Joyce Chai and
                  Natalie Schluter and
                  Joel R. Tetreault},
  title        = {GoEmotions: {A} Dataset of Fine-Grained Emotions},
  booktitle    = {Proceedings of the 58th Annual Meeting of the Association for Computational
                  Linguistics, {ACL} 2020, Online, July 5-10, 2020},
  pages        = {4040--4054},
  publisher    = {Association for Computational Linguistics},
  year         = {2020},
  url          = {https://doi.org/10.18653/v1/2020.acl-main.372},
  doi          = {10.18653/V1/2020.ACL-MAIN.372},
  bibsource    = {dblp computer science bibliography, https://dblp.org}
}

@inproceedings{DBLP:conf/emnlp/LeonardelliMAGT21,
  author       = {Elisa Leonardelli and
                  Stefano Menini and
                  Alessio Palmero Aprosio and
                  Marco Guerini and
                  Sara Tonelli},
  editor       = {Marie{-}Francine Moens and
                  Xuanjing Huang and
                  Lucia Specia and
                  Scott Wen{-}tau Yih},
  title        = {Agreeing to Disagree: Annotating Offensive Language Datasets with
                  Annotators' Disagreement},
  booktitle    = {Proceedings of the 2021 Conference on Empirical Methods in Natural
                  Language Processing, {EMNLP} 2021, Virtual Event / Punta Cana, Dominican
                  Republic, 7-11 November, 2021},
  pages        = {10528--10539},
  publisher    = {Association for Computational Linguistics},
  year         = {2021},
  url          = {https://doi.org/10.18653/v1/2021.emnlp-main.822},
  doi          = {10.18653/V1/2021.EMNLP-MAIN.822},
  bibsource    = {dblp computer science bibliography, https://dblp.org}
}

@inproceedings{DBLP:conf/naacl/JangF24,
  author       = {Hyewon Jang and
                  Diego Frassinelli},
  editor       = {Kevin Duh and
                  Helena G{\'{o}}mez{-}Adorno and
                  Steven Bethard},
  title        = {Generalizable Sarcasm Detection is Just Around the Corner, of Course!},
  booktitle    = {Proceedings of the 2024 Conference of the North American Chapter of
                  the Association for Computational Linguistics: Human Language Technologies
                  (Volume 1: Long Papers), {NAACL} 2024, Mexico City, Mexico, June 16-21,
                  2024},
  pages        = {4238--4249},
  publisher    = {Association for Computational Linguistics},
  year         = {2024},
  url          = {https://doi.org/10.18653/v1/2024.naacl-long.238},
  doi          = {10.18653/V1/2024.NAACL-LONG.238},
  bibsource    = {dblp computer science bibliography, https://dblp.org}
}

@article{DBLP:journals/corr/abs-2510-08460,
  author       = {Elisa Leonardelli and
                  Silvia Casola and
                  Siyao Peng and
                  Giulia Rizzi and
                  Valerio Basile and
                  Elisabetta Fersini and
                  Diego Frassinelli and
                  Hyewon Jang and
                  Maja Pavlovic and
                  Barbara Plank and
                  Massimo Poesio},
  title        = {LeWiDi-2025 at NLPerspectives: The Third Edition of the Learning with
                  Disagreements Shared Task},
  journal      = {CoRR},
  volume       = {abs/2510.08460},
  year         = {2025},
  url          = {https://doi.org/10.48550/arXiv.2510.08460},
  doi          = {10.48550/ARXIV.2510.08460},
  eprinttype   = {arXiv},
  eprint       = {2510.08460},
  bibsource    = {dblp computer science bibliography, https://dblp.org}
}

@article{DBLP:journals/corr/abs-2410-15326,
  author       = {Hsiu{-}Yuan Huang and
                  Yutong Yang and
                  Zhaoxi Zhang and
                  Sanwoo Lee and
                  Yunfang Wu},
  title        = {A Survey of Uncertainty Estimation in LLMs: Theory Meets Practice},
  journal      = {CoRR},
  volume       = {abs/2410.15326},
  year         = {2024},
  url          = {https://doi.org/10.48550/arXiv.2410.15326},
  doi          = {10.48550/ARXIV.2410.15326},
  eprinttype   = {arXiv},
  eprint       = {2410.15326},
  bibsource    = {dblp computer science bibliography, https://dblp.org}
}

@article{DBLP:journals/csur/ShorinwaMLRM26,
  author       = {Ola Shorinwa and
                  Zhiting Mei and
                  Justin Lidard and
                  Allen Z. Ren and
                  Anirudha Majumdar},
  title        = {A Survey on Uncertainty Quantification of Large Language Models: Taxonomy,
                  Open Research Challenges, and Future Directions},
  journal      = {{ACM} Comput. Surv.},
  volume       = {58},
  number       = {3},
  pages        = {63:1--63:38},
  year         = {2026},
  url          = {https://doi.org/10.1145/3744238},
  doi          = {10.1145/3744238},
  bibsource    = {dblp computer science bibliography, https://dblp.org}
}

@article{DBLP:journals/ijmms/XuSL25,
  author       = {Zhengtao Xu and
                  Tianqi Song and
                  Yi{-}Chieh Lee},
  title        = {Confronting verbalized uncertainty: Understanding how LLM's verbalized
                  uncertainty influences users in AI-assisted decision-making},
  journal      = {Int. J. Hum. Comput. Stud.},
  volume       = {197},
  pages        = {103455},
  year         = {2025},
  url          = {https://doi.org/10.1016/j.ijhcs.2025.103455},
  doi          = {10.1016/J.IJHCS.2025.103455},
  bibsource    = {dblp computer science bibliography, https://dblp.org}
}

@inproceedings{DBLP:conf/acl/FadeevaRSPLMTKP24,
  author       = {Ekaterina Fadeeva and
                  Aleksandr Rubashevskii and
                  Artem Shelmanov and
                  Sergey Petrakov and
                  Haonan Li and
                  Hamdy Mubarak and
                  Evgenii Tsymbalov and
                  Gleb Kuzmin and
                  Alexander Panchenko and
                  Timothy Baldwin and
                  Preslav Nakov and
                  Maxim Panov},
  editor       = {Lun{-}Wei Ku and
                  Andre Martins and
                  Vivek Srikumar},
  title        = {Fact-Checking the Output of Large Language Models via Token-Level
                  Uncertainty Quantification},
  booktitle    = {Findings of the Association for Computational Linguistics, {ACL} 2024,
                  Bangkok, Thailand and virtual meeting, August 11-16, 2024},
  series       = {Findings of {ACL}},
  pages        = {9367--9385},
  publisher    = {Association for Computational Linguistics},
  year         = {2024},
  url          = {https://doi.org/10.18653/v1/2024.findings-acl.558},
  doi          = {10.18653/V1/2024.FINDINGS-ACL.558},
  bibsource    = {dblp computer science bibliography, https://dblp.org}
}

@misc{zhang2026tokurtokenleveluncertaintyestimation,
      title={TokUR: Token-Level Uncertainty Estimation for Large Language Model Reasoning}, 
      author={Tunyu Zhang and Haizhou Shi and Yibin Wang and Hengyi Wang and Xiaoxiao He and Zhuowei Li and Haoxian Chen and Ligong Han and Kai Xu and Huan Zhang and Dimitris Metaxas and Hao Wang},
      year={2026},
      eprint={2505.11737},
      archivePrefix={arXiv},
      primaryClass={cs.LG},
      url={https://arxiv.org/abs/2505.11737}, 
}


\end{document}